\documentclass[lettersize,journal]{IEEEtran}
\usepackage{amsmath,amsfonts}
\usepackage{algorithmic}
\usepackage{algorithm}
\usepackage{array}
\usepackage[caption=false,font=normalsize,labelfont=sf,textfont=sf]{subfig}
\usepackage{textcomp}
\usepackage{stfloats}
\usepackage{url}
\usepackage{verbatim}
\usepackage{graphicx}
\usepackage{cite}
\usepackage{hyperref} 
\usepackage{orcidlink}
\usepackage{xcolor} 
\usepackage{cleveref}
\usepackage{caption}
\usepackage{subcaption}
\usepackage{booktabs} 
\usepackage{multirow} 
\usepackage{siunitx}  
\usepackage{array}
\usepackage{colortbl} 
\usepackage{amssymb}
\begin{document}

\title{MCLRL: A Multi-Domain Contrastive Learning with Reinforcement Learning Framework for Few-Shot Modulation Recognition}

\author{Dongwei Xu\orcidlink{0000-0003-2693-922X},~\IEEEmembership{Member, IEEE}, Yutao Zhu\orcidlink{0009-0005-7154-1917}, Yao Lu\orcidlink{0000-0003-0655-7814}, Youpeng Feng\orcidlink{0009-0004-2379-025X}, Yun Lin\orcidlink{0000-0003-1379-9301},~\IEEEmembership{Member, IEEE}, Qi Xuan\orcidlink{0000-0002-6320-7012},~\IEEEmembership{Senior Member,~IEEE}
\thanks{This work was partially supported by the Key R\&D Program of Zhejiang under Grant 2022C01018 and by the National Natural Science Foundation of China under Grant U21B2001 and Grant 61973273. (Corresponding author: Qi Xuan)}
\thanks{Yutao Zhu and Youpeng Feng are with the Institute of Cyberspace Security, College of Information Engineering, Zhejiang University of Technology, Hangzhou, China, also with the Binjiang Institute of Artificial Intelligence, Zhejiang University of Technology, Hangzhou 310056, China (e-mail:  zhuyutao629@gmail.com, fengypmon@gmail.com.}
\thanks{Yun Lin is with the College of Information and Communication Engineering, Harbin Engineering University, Harbin, China (e-mail: linyun@hrbeu.edu.cn).}
\thanks{Yao Lu, Dongwei Xu and Qi Xuan are with the Institute of Cyberspace Security, Zhejiang University of Technology, Hangzhou 310023, China, and also with the Binjiang Institute of Artificial Intelligence, Zhejiang University of Technology, Hangzhou 310056, China (e-mail: yaolu.zjut@gmail.com, dongweixu@zjut.edu.cn, xuanqi@zjut.edu.cn).}
}

\markboth{Journal of \LaTeX\ Class Files,~Vol.~14, No.~8, August~2021}%
{Shell \MakeLowercase{\textit{et al.}}: A Sample Article Using IEEEtran.cls for IEEE Journals}

\IEEEpubid{0000--0000/00\$00.00~\copyright~2021 IEEE}

\maketitle
\begin{abstract}
With the rapid advancements in wireless communication technology, automatic modulation recognition (AMR) plays a critical role in ensuring communication security and reliability. However, numerous challenges, including higher performance demands, difficulty in data acquisition under specific scenarios, limited sample size, and low-quality labeled data, hinder its development. Few-shot learning (FSL) offers an effective solution by enabling models to achieve satisfactory performance with only a limited number of labeled samples. While most FSL techniques are applied in the field of computer vision, they are not directly applicable to wireless signal processing. This study does not propose a new FSL-specific signal model but introduces a framework called MCLRL. This framework combines multi-domain contrastive learning with reinforcement learning. Multi-domain representations of signals enhance feature richness, while integrating contrastive learning and reinforcement learning architectures enables the extraction of deep features for classification. In downstream tasks, the model achieves excellent performance using only a few samples and minimal training cycles. Experimental results show that the MCLRL framework effectively extracts key features from signals, performs well in FSL tasks, and maintains flexibility in signal model selection.
\end{abstract}


\section{Introduction}
\IEEEPARstart{T}{he} advent of wireless communication technology, with its rapid development and widespread application, has led to increasingly complex radio signals~\cite{tang2018digital}. Early radio signal recognition primarily relied on specialized equipment and manual operations. However, as the signal load has increased, the spectrum range has expanded, and accuracy requirements have grown. It has become clear that traditional methods, relying on manual effort and basic equipment, are no longer sufficient to meet the demands of modern society~\cite{ding2018amateur}.

As an emerging technology, deep learning has rapidly advanced in recent years, achieving significant results in areas such as image classification~\cite{deng2014deep,he2016deep}, speech recognition~\cite{wang2019overview,malik2021automatic}, and natural language processing~\cite{chowdhary2020natural,khurana2023natural}. The computational power and pattern recognition capabilities of deep learning offer new possibilities for solving radio signal recognition problems. As research has advanced, deep learning algorithms have been increasingly applied to key areas of radio signal systems~\cite{tu2018semi,lu2024generic,wang2019data,ya2022large}, achieving significant results in the field of AMR. However, the application of deep learning typically requires large amounts of labeled data, and the collection and labeling process demands substantial human and material resources, limiting the widespread use of deep learning technologies in complex communication environments. Consequently, traditional supervised AMR methods only apply to scenarios with abundant labeled data~\cite{qi2020small}. However, in practical applications, data can be easily collected, but the high cost of labeling remains a common issue~\cite{zhu2005semi}. Effectively training an encoder with limited labeled data is a core challenge that must be addressed in the AMR field.
\IEEEpubidadjcol
To address the issue of label scarcity, many enhancement techniques and methods based on meta-learning and metric learning have emerged. These include AFHN~\cite{li2020adversarial}, MAML~\cite{finn2017model}, and ProtoNet~\cite{snell2017prototypical}. Typically, these meta-learning and metric learning methods learn efficient base models from large datasets and adapt them to few-shot tasks using distance metrics or gradient optimization. However, in radio signal processing, the availability of base classes is often limited, and the direct application of these methods may lead to inadequate model generalization. In FSL scenarios for AMR tasks, using a large number of unlabeled samples from target datasets for contrastive learning has proven to be an effective solution. In recent years, contrastive learning, an emerging self-supervised learning technique~\cite{fan2021unsupervised}, has become a powerful method for representing the features of unlabeled signals by exploring implicit supervisory information in large-scale datasets. For instance, Liu et al.\cite{liu2021self} were the first to introduce contrastive learning into AMR tasks, using the SimCLR\cite{chen2020simple} training method and applying a single rotation transformation to the augmented signal data. Later, Davaslioglu et al.\cite{davaslioglu2022self} proposed a method using MoCo\cite{he2020momentum} to learn feature representations of unlabeled signals.

Despite advancements in the field, significant challenges remain in meeting the demand for efficient feature representation of unlabeled modulation signal, particularly due to the following technical difficulties.

\begin{itemize}
\item[$\bullet$] The effectiveness of self-supervised contrastive learning methods depends on exploring hidden features in unlabeled signals through data augmentation. However, various interferences in real-world electromagnetic environments often cause severe distortion and distribution shifts in received signals. This negatively impacts the classification tasks of unlabeled signals using contrastive learning, limiting the full use of hidden features. As a result, traditional data augmentation methods and parameters often need to be adjusted and optimized for different environments.
\item[$\bullet$] A single data representation space is often not enough for encoders to fully extract diverse hidden features from unlabeled signals. This limitation prevents some traditional AMR methods in few-shot scenarios from capturing sufficient feature information from a single representation domain, which affects the performance of downstream classification tasks.
\end{itemize}

To address these challenges, we propose a pretraining framework (MCLRL) based on reinforcement learning and multi-representation domain collaborative contrastive learning to achieve efficient feature representation for unlabeled modulation signal. Specifically, the idea is to use reinforcement learning to replace the manual selection of data augmentation methods and intensities, allowing the extraction of more useful features tailored to different datasets and models. The diverse data representations of modulation signal can be utilized through multi-domain joint training to extract rich feature information~\cite{qi2020automatic}. Based on this concept, we integrate three representations—IQ signals, frequency domain, and constellation diagrams—and design a reinforcement learning-based optimal selection strategy to improve the application of contrastive learning for modulation signal feature representation. The proposed MCLRL framework consists of three stages: 1) Encoder pretraining, 2) Feature fusion and classifier fine-tuning, and 3) Downstream task evaluation. In the encoder pretraining stage, contrastive learning uses mutual supervision across the three representation domains, with reinforcement learning selecting the optimal data augmentation intensity. Features extracted by the encoder are normalized through a projection head and integrated using an intra-domain and inter-domain joint loss function. In the next stage, the encoder parameters are frozen, and a lightweight feature fusion module is used to extract critical features from the three representation domains. These features are then passed into a classifier for classification. In the downstream task evaluation stage, all module parameters are frozen, and performance is tested.

The primary contributions of this paper are as follows:
\begin{itemize}
\item[$\bullet$] A new few-shot AMR framework, MCLRL, is introduced. This framework consists of three components, designed to more effectively evaluate the performance of AMR models in scenarios with large amounts of unlabeled data and limited labeled data. Unlike traditional contrastive learning methods based solely on IQ sequences, MCLRL incorporates frequency domain and constellation diagram information, fully exploring the diverse feature representations of modulation signal.
\item[$\bullet$] A reinforcement learning-based contrastive learning pretraining encoder method is designed. The proposed method employs an optimal selection strategy using reinforcement learning within a continuous action space to select the most suitable data augmentation methods. This approach is more effective than manually adjusting augmentation parameters, as it better explores hidden features in modulation signal and enhances the distinguishability of these features.
\item[$\bullet$] An intra-domain and inter-domain joint loss function is designed. This loss function allows the three different representation domains of modulation signal to learn from and supervise each other in the same high-dimensional space. It also expands the positive and negative sample pairs in contrastive learning, leading to more effective separation in that space. A lightweight feature fusion module for few-shot downstream tasks is designed. The proposed module has a small parameter size and enables classifier fine-tuning with minimal sample sizes, effectively avoiding overfitting in few-shot scenarios.
\item[$\bullet$] Our MCLRL framework achieves an accuracy improvement of 3\% to 15\% over traditional image contrastive learning methods, time-series contrastive learning methods, and modulation signal FSL methods, on the RML2016.10a and Sig2019-12 datasets. This highlights the superiority of our approach.
\end{itemize}

The remainder of this article is structured as follows. Section~\ref{sec:Related Works} discusses related work on AMR and contrastive learning. Section~\ref{sec:PROBLEM FORMULATION} introduces the problem statement and background knowledge. Section~\ref{sec:MCLRL Framework} provides a detailed description of the components of MCLRL. Section~\ref{sec:Experiments} presents the experimental design and results of MCLRL on different datasets, and Section~\ref{sec:Conclusion} concludes the paper.

\section{Related Works}
\label{sec:Related Works}
\subsection{DL-Based Automatic Modulation Recognition}
AMR is a key technology in communications signal processing~\cite{rajendran2018deep} that aims to automatically identify the modulation type of received signals. This technology has been widely applied in communication fields such as interference detection~\cite{van2008electromagnetic,xie2019localization}, spectrum sensing~\cite{long2015fully,wang2018cyber}, and electronic countermeasures~\cite{maglaras2014intrusion,butt2018hybrid}. With the rapid development of wireless communication technology, modern wireless communication scenarios have become more complex and the data volume has increased dramatically. Traditional experience-based manual feature extraction methods can no longer meet the requirements. Deep learning, with its strong feature extraction capabilities and advantages in processing large data sets, has gradually become the mainstream method in the AMR field ~\cite{zhang2021radio}.

O'Shea et al.~\cite{o2016convolutional} pioneered the approach of treating IQ radio signals as single-channel images of width 2, using a simple 2D convolutional network and fully connected layers for feature extraction and classification. They also published two public modulation signal datasets, laying the foundation for subsequent research. Chen et al.~\cite{chen2021signet} further developed a framework that combined the signal-to-matrix (S2M) operator with a residual network, achieving collaborative optimization between input images and deep models. Although high recognition accuracy was achieved, the computational resource requirements were relatively demanding. Xu et al.~\cite{xu2020spatiotemporal} proposed a novel 3D deep learning framework combining convolutional neural networks (CNN) and bidirectional long short-term memory (BiLSTM) models to extract joint features from I/Q symbols, successfully addressing the long-term dependency issue.

In addition, some researchers have attempted to combine deep learning with expert knowledge. For example, Ding et al.~\cite{ding2022data} developed a data- and knowledge-driven modulation recognition method that significantly improved recognition performance in complex scenarios. Zhang et al.~\cite{zhang2021efficient,zhang2018modulation} proposed novel modulation recognition models based on parameter estimation and time-frequency distributions. At the framework level, Qi et al.~\cite{qi2020automatic} designed an efficient framework by combining deep residual networks with multi-modal information. These methods perform well in data-rich scenarios, but their performance drops significantly when sample sizes are small, limiting their practical value in real-world communication environments.

\subsection{Few-Shot Learning}
In modulation signal recognition, FSL is a challenging but crucial task, especially in cases of data scarcity or limited labeled data. Currently, FSL has made significant progress in areas such as image~\cite{zhang2023learning}, text~\cite{geng2019induction,yan2018few,feng2021survey}, and audio~\cite{heggan2022metaaudio,wang2021few}. However, research in the field of modulation signal is still relatively sparse. Existing FSL methods can be broadly categorized into four types. 1) Methods based on prior knowledge: These methods introduce expert or domain knowledge to analyze the features, frequency, and other aspects of the signal to enable effective detection under few-shot conditions. 2) Data augmentation methods: These methods expand the sample size using data augmentation techniques, such as Generative Adversarial Networks (GANs)~\cite{tang2018digital}. Although GANs can generate high-quality samples, they face problems such as unstable training and limited generalization, particularly in few-shot scenarios where they are less applicable. 3) Methods based on model optimization: These approaches use a combination of pretraining and fine-tuning to improve model performance in few-shot scenarios through transfer learning. In addition, designing lightweight architectures specifically for few-shot problems is also an effective strategy. 4) Methods based on algorithm optimization: Methods such as meta-learning~\cite{finn2017model} optimize learning algorithms to adapt to few-shot scenarios. Although theoretically attractive, these methods are often limited in practical applications due to the similarity of source datasets.

\subsection{Self-Supervised Contrast Learning}
Self-supervised learning, an important branch of unsupervised learning, learns effective representations from large amounts of unlabeled data by designing surrogate tasks. In recent years, it has shown outstanding performance in areas such as natural language processing~\cite{devlin2018bert}, computer vision~\cite{he2022masked}, and audio processing~\cite{niizumi2021byol}. Contrastive learning is one of the core methods of self-supervised learning, where it encourages the model to learn deep semantic features from the data by contrasting the similarity between positive and negative sample pairs.

Classical methods such as SimCLR~\cite{chen2020simple} and MoCoV2~\cite{chen2020improved} optimize contrast loss by using large positive and negative samples and dynamic queues, respectively, while BYOL~\cite{grill2020bootstrap} and SwAV~\cite{zhu2020swav} improve performance by removing negative samples or introducing clustering ideas. In the time series domain, TS-TCC~\cite{eldele2021time} develops an unsupervised representation learning framework for sleep classification tasks using temporal and contextual contrast. Liu et al.~\cite{liu2021self} applied SimCLR to electromagnetic modulation classification and designed a semi-supervised framework. In addition, some studies focus on individual recognition and radio frequency fingerprinting, such as the SA2SEI framework~\cite{liu2023overcoming}, which overcomes data limitations by combining self-supervised learning and adversarial augmentation.

Despite the significant innovations of the above methods in time series, modulation signal modulation detection still faces challenges, such as high-dimensional data and the need for large labeled samples. Designing a framework that balances adaptability to complex scenarios with efficient use of data samples is an important direction for future research in this area.

\section{Background Knowledge and Problem Formulation}
\label{sec:PROBLEM FORMULATION}
\subsection{Background Knowledge}
In modulation detection tasks, the received radio signals are typically in the time domain (modulation signal), and $x(t)$ is expressed as:
\begin{equation}
\label{eq1}
x(t)=\mathcal{S}(t)*h(t)\exp\left(j2\pi\Delta ft+\psi_0\right)+\mathrm{n}(t),
\end{equation}
where $*$ represents the convolution operation, $\mathcal{S}(t)$ denotes the modulation signal, $h(t)$ is the impulse response of the wireless channel, $\Delta f$ is the carrier frequency offset, $\psi_0$ is the initial phase, and $\mathrm{n}(t)$ represents the environmental noise. Radio signals are typically modulated, so the signal can be represented as two channels:
\begin{equation}
\label{eq2}
x(t)=I(t)+jQ(t).
\end{equation}
For any signal of length $N$, after sampling, the time domain representation of the signal can be expressed as:
\begin{equation}
\label{eq3}
x=\begin{bmatrix}I \\Q\end{bmatrix}, \quad \text{where} \begin{cases}
I=[i_1,i_2,...,i_N] \\
Q=[q_1,q_2,...,q_N] & \end{cases}.
\end{equation}
Therefore, for each modulation type label $y$, $(x, y)$ can be considered as a pair of independent variables $x$ and dependent variable $y$. The goal is to have the neural network model the mapping relationship between $x$ and $y$, denoted as $f(\bullet)$. This mapping can then be used by the neural network to model and perform the modulation signal recognition task.
\subsection{Problem Formulation}
The scenario addressed in this paper is small sample modulation signal recognition, specifically simulating common problems in modulation signal labeling, such as difficulties in labeling, low efficiency, and the high cost of manual labeling. There is a large amount of unlabeled data that needs to be labeled and a minimal amount of labeled data, with all categories represented. Our goal is to train a model that can label similar data in the future. Therefore, the process is mainly divided into two stages: 1) Training phase: The model is trained on a $base$ set consisting of a large amount of unlabeled data covering all classes. It is then fine-tuned on a $support$ set consisting of a small amount of labeled data with the same classes as the $base$ set. 2) Testing phase: Performance testing is carried out on a $query$ set consisting of data belonging to the same classes as the $base$ and $support$ sets, but without overlapping with them.

To be more precise, we define the $base$ set as $\mathcal{D}_b=\{x^b, y^b\}$, where $x^b$ is the quadrature modulation signal segment and the corresponding label $y^b\subset\mathbb{R}^{\mathcal{D}_{train}}$ belongs to a total of $\mathcal{D}_{train}$ all classes. The $support$ set is denoted as $\mathcal{D}_s=\{x^s, y^s\}, y^s\subset\mathbb{R}^{\mathcal{D}_{train}}$, where $\mathcal{D}_{train}$ is the remaining samples except the $base$ set, and each class contains $N$ samples. The $query$ set is denoted as $\mathcal{D}_q=\{x^q, y^q\}, y^q\subset\mathbb{R}^{\mathcal{D}_{test}}$, where $\mathcal{D}_{test}$ and $\mathcal{D}_{train}$ have the same category. It is important to note that the $base$ set, $support$ set, and $query$ set categories are all the same, while the $base$ and $support$ sets form the $train$ set and the $query$ sets form the $test$ set (i.e., $\mathcal{D}_b\cup \mathcal{D}_s=\mathcal{D}_{train}$, and $\mathcal{D}_q=\mathcal{D}_{test}$). Therefore, the objectives set in this paper can be expressed as follows:
\begin{equation}
\label{eq4}
\min(\epsilon_{\mathrm{error}})=\mathbb{E}_{(x^q,y^q)\sim \mathcal{D}_q}[f^{\prime}(x^q)\neq y^q],
\end{equation}
where $\epsilon_{\mathrm{error}}$ denotes the target error on $query$ set and the $f^{\prime}$ represents the trained models using $base$ and $support$ set.

\section{MCLRL Framework}
\label{sec:MCLRL Framework}
\begin{figure*}[!t]
\centering
\includegraphics[width=\textwidth, clip, trim=0.15in 0.15in 0.15in 0.15in]{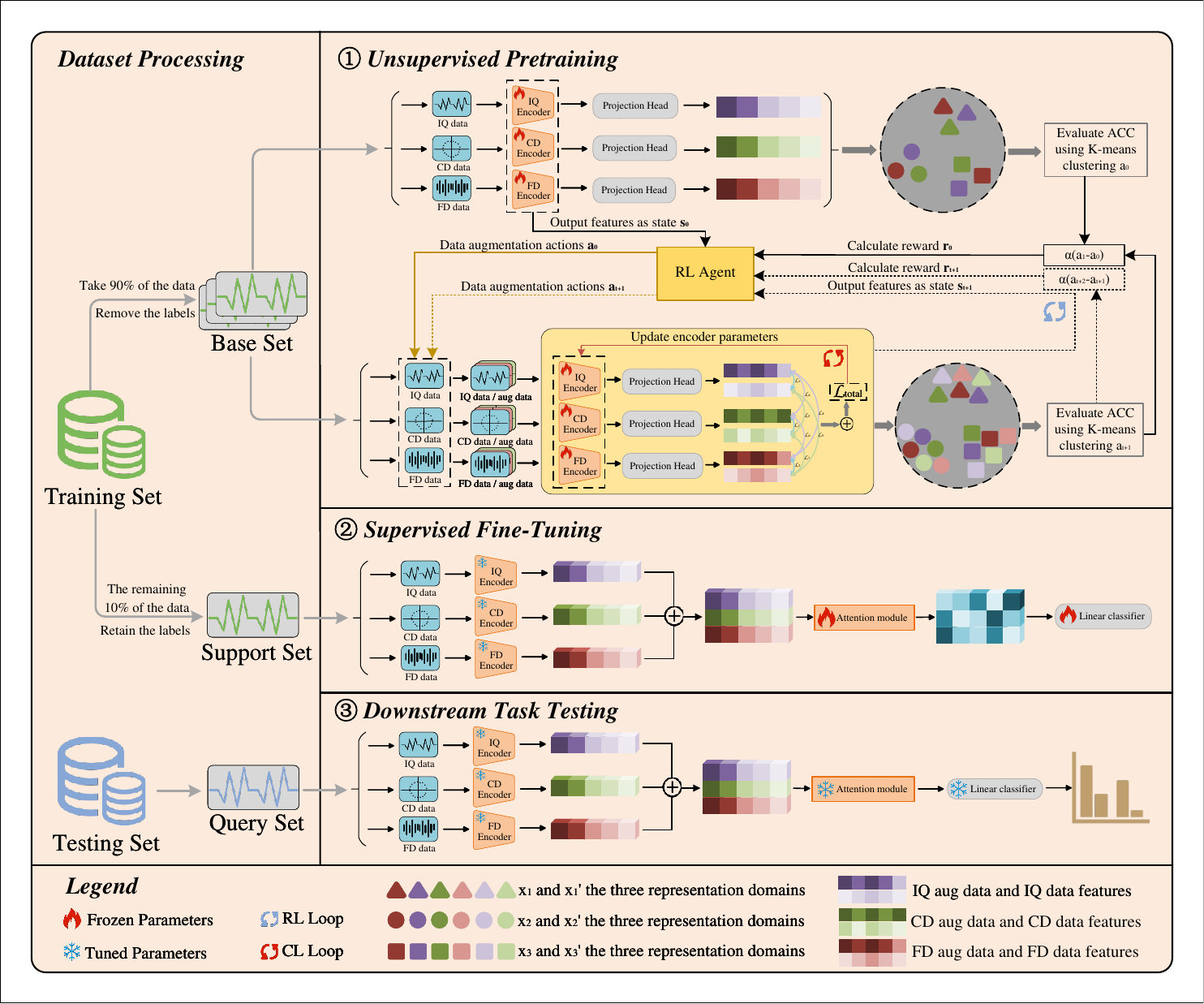}
\caption{Multi-domain contrast learning MCLRL framework. The MCLRL framework can be divided into four parts: data processing, unsupervised pretraining, supervised fine-tuning, and downstream task testing.}
\label{fig_2}
\end{figure*}

This section introduces the proposed Multi-domain Contrastive Learning (MCLRL) framework for few-shot modulation signal recognition, as shown in Fig.~\ref{fig_2}. The MCLRL framework consists of three main parts: unsupervised pretraining, supervised fine-tuning, and downstream task testing. In the data processing step, a complete dataset is divided into training and testing sets. The training set is split into a $base$ set (unlabeled) and a $support$ set (labeled), while the testing set constitutes the $query$ set. For each dataset, the IQ data is preserved, while additional transformations generate constellation diagram data and frequency domain data, forming three different representation domains. Each of these three domains undergoes five different data augmentation methods (noise addition, time shifting, scaling, dropout, and interpolation), from which reinforcement learning will select. In the unsupervised pretraining phase, contrastive learning is combined with reinforcement learning for training. The 'state' in reinforcement learning is defined as the concatenated features output by the encoder, the 'action' refers to selecting the type and extent of data augmentation, and the 'reward' is the improvement in classification accuracy derived from K-means clustering~\cite{ahmed2020k}. In the first round of reinforcement learning training, both the 'action' and 'reward' are initially set to 0, and only the 'state' is extracted so the agent can learn and generate actions. In subsequent rounds, the three different encoders perform contrastive learning based on the given 'actions', and after each round, performance is evaluated to calculate the 'reward'. In the supervised fine-tuning phase, the encoder parameters are frozen, and the $support$ set is used to fine-tune the attention module and linear classifier. In the downstream task testing phase, all module parameters are frozen, and the final classification performance is evaluated using the $query$ set.

\subsection{Data Processing and Augmentation}
After dividing the data into the $base$ set, $support$ set, and $query$ set, further transformation into different representation domains is required. First, the I (in-phase component) and Q (quadrature component) of the IQ signals are assembled into complex signals. Then, the Fast Fourier Transform (FFT)~\cite{rao2011fast} is utilized to compute the representation of the signals in the frequency domain, generating both the magnitude spectrum and the phase spectrum. Assume the input signal is $IQ=[I, Q]$, with a signal length of $N$. The time domain signal can be expressed as~\cref{eq3}, where the signal is converted from the time domain to the frequency domain using FFT:
\begin{equation}
\label{eq5}
x[k]=\sum_{n=0}^{N-1}x[n]e^{-j2\pi kn/N},\quad k=1,2,\ldots,N,
\end{equation}
where $x[k]$ represents the frequency domain representation of the signal. The magnitude spectrum and phase spectrum are then computed and combined to form a tensor $\mathbf{R}$ with a shape of $[2, N]$:
\begin{equation}
\label{eq6}
|x[k]|=\sqrt{\mathrm{Re}(x[k])^2+\mathrm{Im}(x[k])^2},
\end{equation}
\begin{equation}
\label{eq7}
\angle{x[k]}=\arctan\left(\frac{\mathrm{Im}(x[k])}{\mathrm{Re}(x[k])}\right),
\end{equation}
\begin{equation}
\label{eq8}
\mathbf{R}=
\begin{bmatrix}
\mathrm{Re}(x) \\
\mathrm{Im}(x)
\end{bmatrix}.
\end{equation}
The above process implements the transformation of the signal from the time domain to the frequency domain, providing a foundation for subsequent feature extraction and analysis. In the frequency domain, the magnitude spectrum $|x[k]|$ represents the energy distribution of the signal, while the phase spectrum $\angle{x[k]}$ contains the time-related information of the signal. This frequency domain representation is crucial in communication signal processing and pattern recognition.

\begin{figure}[t]
\centering
\includegraphics[width=3.0in, clip, trim=0.5in 0.25in 0.25in 0.25in]{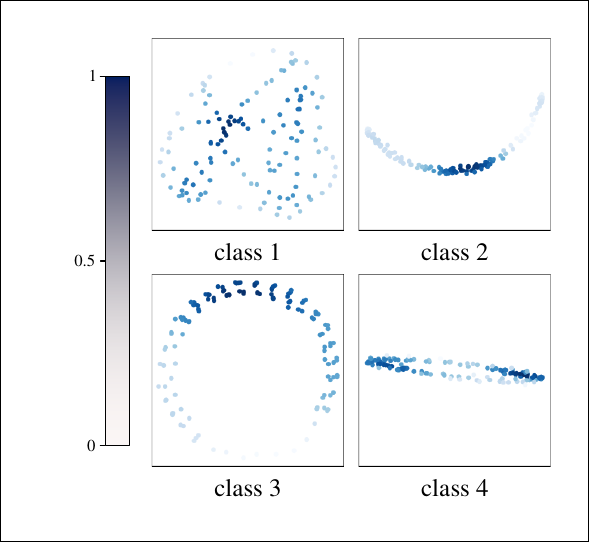}
\caption{The heatmap-based constellation diagrams for four different categories in the RML2016.10a dataset are shown below. In these diagrams, the color darkens as the points become more concentrated.}
\label{fig_1}
\end{figure}

Next, we need to convert the IQ signals into constellation diagrams. Each sample $x[n]$ corresponds to a point on the complex plane, with the in-phase component $I[n]$ on the horizontal axis and the quadrature component $Q[n]$ on the vertical axis. By plotting all samples on the complex plane, the resulting set of points forms the constellation diagram. The method of drawing the constellation diagram is simple; however, for the same batch of data, it is unclear whether each class corresponds to an intuitive shape because the shape of the constellation diagram is influenced by the signal's sampled frequency domain~\cite{liu2014multidimensional}. As a result, even samples from the same class may have different or irregular constellation shapes. To mitigate this issue, we use heatmap-based constellation diagrams, enhancing the color variation to enrich their semantic information, as shown in Fig.~\ref{fig_1}. Regions with denser points are colored darker, while regions with sparser points are colored lighter.

After converting the IQ signals into the frequency domain and constellation diagrams, data augmentation is applied to each of the three representation domains. For both time domain IQ signals and frequency domain signals, the same data augmentation methods are used because of their identical length and tensor dimensions. This ensures consistency across the representations in different domains. Whether the signal is augmented in the time or frequency domain, its feature representations should remain similar, helping the model learn generalizable features. For constellation diagrams, since they are in image form, the augmentation methods must be chosen as consistently as possible to maintain uniformity.

Taking IQ signals as an example, five common augmentation methods are used: noise addition, time shifting, scaling, random dropout, and interpolation. Adding noise augments the model to handle signal interference that may occur in real-world communication environments, thereby improving its robustness. Given the original signal $x[n]$, the noise addition process can be represented by the following formula:
\begin{equation}
\label{eq9}
x_\mathrm{noisy}[n]=x[n]+\sigma\cdot\epsilon[n],
\end{equation}
where $x_\mathrm{noisy}[n]$ is the signal with added noise, $\epsilon[n]\sim\mathcal{N}(0,1)$ represents the standard normal distribution noise, and $\sigma$ is the standard deviation of the noise, controlling the amplitude of the noise. Time shifting simulates potential delays or synchronization errors that may occur during signal transmission. time shifting is achieved by displacing the signal along the time axis. This process can be represented as:
\begin{equation}
\label{eq10}
x_\mathrm{shifted}[n]=x[(n+\Delta){\mathrm{mod}}N],
\end{equation}
where $x_\mathrm{shifted}[n]$ is the signal after time shifting, $\Delta$ is the randomly generated shift value, which ranges from $[-\Delta_{\max},\Delta_{\max}]$, and $N$ is the total length of the signal. Signal scaling is used to simulate amplitude variations that may occur due to signal amplification or attenuation. the scaling operation adjusts the amplitude of the signal by multiplying it by a random scaling factor. The formula is:
\begin{equation}
\label{eq11}
x_\mathrm{scaled}[n]=\alpha\cdot{x[n]},
\end{equation}
where $\alpha\in[\alpha_{\min},\alpha_{\max}]$ is a randomly generated scaling factor that controls the range of amplitude variation of the signal. Random dropout aims to simulate data loss or noise interference during the signal transmission process. This method generates sparse signals by discarding a portion of the signal samples with a certain probability. the dropout operation can be formalized as:
\begin{equation}
\label{eq12}
x_\mathrm{dropped}[n]=x[n]\cdot\mathbf{m}[n],
\end{equation}
where $\mathbf{m}[n]$ is a randomly generated mask matrix, with elements being binary random variables that satisfy $\mathbf{m}[n]\sim\mathrm{Bernoulli}(1-p)$, where $p$ is the probability of dropout. The interpolation method is used to adjust the time resolution of the signal, simulating the effects of different sampling rates or time intervals on the signal. The interpolation operation alters the signal's time length by increasing or decreasing the number of sampling points and is expressed as:
\begin{equation}
\label{eq13}
x_\mathrm{interpolated}{[n]}=f(x[n],\gamma),
\end{equation}
where $\gamma$ is the interpolation factor, which controls the length of the signal after interpolation. Typically, $\gamma>1$ represents signal length extension, while $\gamma<1$ represents signal length shortening.

\subsection{Unsupervised pretraining}
\textit{1) Multi-domain Contrastive Learning Loop:} Formally, for a batch of signal samples $\mathbf{x}_i$, where $i\in[1,B]$ and $B$ represents the batch size, the time domain, frequency domain, and constellation diagram domain can be represented as $\mathbf{x}_i^T$, $\mathbf{x}_i^F$, and $\mathbf{x}_i^C$ after multi-representation domain transformation. For each sample, one of the five data augmentation methods is randomly selected. The augmented samples in the three representation domains can then be denoted as $\tilde{\mathbf{x}}_i^T$, $\tilde{\mathbf{x}}_i^F$, and $\tilde{\mathbf{x}}_i^C$. Consider the time domain, after encoder and projection head, it is projected into the representation space to obtain $\mathbf{z}_i^{T}=g_\theta(f_\theta(\mathbf{x}_i^T))$ and $\tilde{\mathbf{z}}_i^{T}=g_\theta(f_\theta(\tilde{\mathbf{x}}_i^T))$. The same can be said for the remaining two representation fields $\mathbf{z}_i^{F}=g_\xi(f_\xi(\mathbf{x}_i^F))$, $\tilde{\mathbf{z}}_i^{F}=g_\xi(f_\xi(\tilde{\mathbf{x}}_i^F))$ and $\mathbf{z}_i^{C}=g_\phi(f_\phi(\mathbf{x}_i^C))$, $\tilde{\mathbf{z}}_i^{C}=g_\phi(f_\phi(\tilde{\mathbf{x}}_i^C))$. After that, the similarity of pairs is calculated on this representation space using cosine similarity. The similarity of two eigenvectors is calculated as:
\begin{equation}
\label{eq14}
\mathrm{sim}\left(\mathbf{z}_i^{T},\tilde{\mathbf{z}}_j^{T}\right)=\frac{\left(\mathbf{z}_i^{T}\right)\left(\tilde{\mathbf{z}}_j^{T}\right)^\top}{\|\mathbf{z}_i^{T}\|_2\|\tilde{\mathbf{z}}_j^{T}\|_2},
\end{equation}
where $i, j\in[1,B]$, $\mathrm{sim}\left(\mathbf{z}_i^{T},\tilde{\mathbf{z}}_j^{T}\right)$ is a similar matrix. The same applies to the frequency domain and the constellation diagram domain, represented as $\mathrm{sim}\left(\mathbf{z}_i^{F},\tilde{\mathbf{z}}_j^{F}\right)$ and $\mathrm{sim}\left(\mathbf{z}_i^{C},\tilde{\mathbf{z}}_j^{C}\right)$, respectively. The intra-domain contrastive loss of the time domain based on cosine similarity can be expressed as:
\begin{equation}
\label{eq15}
\ell_\mathrm{intra}^T=-\mathrm{log}\frac{\exp(\frac{\mathrm{sim}(\mathbf{z}_i^{T},\tilde{\mathbf{z}}_i^{T})}{\tau})}
{\sum_{j=1}^{B-1}\left[\exp(\frac{\mathrm{sim}(\mathbf{z}_i^{T},\mathbf{z}_j^{T})}{\tau})+\exp(\frac{\mathrm{sim}(\mathbf{z}_i^{T},\tilde{\mathbf{z}}_j^{T})}{\tau})\right]},
\end{equation}
where $\tau$ is the temperature parameter, $i \neq j,i\in[1, B], j \in[1,B-1]$. In~\cref{eq15}, the numerator indicates that the original samples and augmented samples have a one-to-one correspondence within a batch of size $B$, making them positive pairs. In the denominator, the first term represents the negative pairs formed by the original sample and other original samples, while the second term represents the negative pairs formed by the original sample and other augmented samples. In contrastive learning, positive pairs should be close in the high-dimensional feature space, while negative pairs should be far apart. The intra-domain contrastive loss for the other two representation domains can be expressed as $\ell_\mathrm{intra}^F$ and $\ell_\mathrm{intra}^C$. The general intra-domain contrastive loss can be formulated as:
\begin{equation}
\label{eq16}
L_\mathrm{intra}=\ell_\mathrm{intra}^T+\ell_\mathrm{intra}^F+\ell_\mathrm{intra}^C.
\end{equation}
On this basis, the introduction of inter-domain contrastive loss can better enable feature learning and supervision between different representation domains. For inter-domain contrastive loss, there are three forms: between original samples, between original and augmented samples, and between augmented samples. However, the inter-domain loss between original and augmented samples involves excessively large feature differences. Such significant differences may lead to reduced inter-class sample distances in high-dimensional space~\cite{zhang2022self}, as also demonstrated in our experiments. Similarly, the inter-domain contrastive loss between the time domain and the frequency domain of an original sample can be expressed as:
\begin{equation}
\label{eq17}
\ell_\mathrm{inter}^{TF}=-\mathrm{log}\frac{\exp(\frac{\mathrm{sim}(\mathbf{z}_i^{T},\mathbf{z}_i^{F})}{\tau})}{\sum_{j=1}^{B-1}\left[\exp(\frac{\mathrm{sim}(\mathbf{z}_i^{T},\mathbf{z}_j^{T})}{\tau})+\exp(\frac{\mathrm{sim}(\mathbf{z}_i^{T},\mathbf{z}_j^{F})}{\tau})\right]}.
\end{equation}
Similarly, we can obtain five other forms of inter-domain contrastive loss, denoted as $\ell_\mathrm{inter}^{TC}$, $\ell_\mathrm{inter}^{FC}$, $\ell_\mathrm{inter}^{\tilde{T}\tilde{F}}$, $\ell_\mathrm{inter}^{\tilde{T}\tilde{C}}$, and $\ell_\mathrm{inter}^{\tilde{F}\tilde{C}}$. Therefore, the total inter-domain contrastive loss can be expressed as:
\begin{equation}
\label{eq18}
L_\mathrm{inter}=\ell_\mathrm{inter}^{TF}+\ell_\mathrm{inter}^{TC}+\ell_\mathrm{inter}^{FC}+\ell_\mathrm{inter}^{\tilde{T}\tilde{F}}+\ell_\mathrm{inter}^{\tilde{T}\tilde{C}}+\ell_\mathrm{inter}^{\tilde{F}\tilde{C}}.
\end{equation}
Therefore, the total contrastive loss for the entire MCLRL framework can be expressed as:
\begin{equation}
\label{eq19}
L_\mathrm{total}=\lambda{L_\mathrm{intra}}+(1-\lambda)L_\mathrm{inter},
\end{equation}
where $\lambda$ is a hyper parameter used to control the weight between intra-domain loss and inter-domain loss.

\textit{2) Reinforcement Learning Loop:} The goal of the reinforcement learning framework is to optimize the data augmentation strategy using the Soft Actor-Critic (SAC) algorithm~\cite{haarnoja2018soft}, thereby enhancing the clustering performance of features output by the encoder. The concatenated features output by the encoder are used to describe the representation distribution of the current sample and serve as the reinforcement learning state $s_{t}$. The data augmentation methods and their corresponding intensities form a continuous action space, where the range is defined by the given augmentation parameter intervals, represented as actions $a_{t}$. The difference in K-means clustering accuracy is used to evaluate the effectiveness of the data augmentation strategy and serves as the reward $r_{t}$, defined as:
\begin{equation}
\label{eq20}
s_t=\mathrm{concat}(f_\theta(\mathbf{x}_i^T),f_\xi(\mathbf{x}_i^F),f_\phi(\mathbf{x}_i^C)).
\end{equation}The action $a_t$ can be defined as:
\begin{equation}
\label{eq21}
a_t=(a_1,a_2,a_3,a_4,a_5),
\end{equation}
where represent five different augmentation methods. To narrow down the exploration range of the action space, we combine the original 15 augmentation methods, where each representation domain has 5 augmentation methods, by multiplying the corresponding degree coefficient with similar augmentation methods. In this way, it is only necessary to explore five different action spaces. Finally, the reward 
$r_t$ can be defined as:
\begin{equation}
\label{eq22}
r_t=\mathrm{acc}_t-\mathrm{acc}_{t-1},
\end{equation}
where $\mathrm{acc}_t$ is the clustering accuracy after applying the current action $a_t$, $\mathrm{acc}_{t-1}$ is the clustering accuracy from the previous round.

Thus, the reinforcement learning process can be represented as a Markov Decision Process (MDP), defined as:
\begin{equation}
\label{eq23}
\mathcal{M}=(\mathcal{S},\mathcal{A},P,R,\gamma),
\end{equation}
where $\mathcal{S}$ is the state space, $\mathcal{A}$ is the action space, $P(s_{t+1}\mid s_t,a_t)$ represents the state transition probability, $R(s_t,a_t)$ is the reward function, and $\gamma$ is the discount factor that balances current and future rewards. Using SAC as the agent, the goal is to learn an optimal policy $\pi(a_t\mid s_t)$ that maximizes the expected cumulative reward:
\begin{equation}
\label{eq24}
J(\pi)=\mathbb{E}_\pi\left[\sum_{t=0}^{E_{rl}}\gamma^tr_t\right],
\end{equation}
where $\gamma\in[0,1)$ is the discount factor, $r_t$ is the instant reward, $epoch_{rl}$ is the reinforcement learning cycle period. The optimization objective of SAC combines the expected reward and the entropy of the policy, defined as:
\begin{equation}
\label{eq25}
J(\pi)=\mathbb{E}_{(s_t,a_t)\sim\pi}\left[Q_\pi(s_t,a_t)+\alpha\mathcal{H}(\pi(\cdot\mid s_t))\right],
\end{equation}
where $Q_\pi(s_t,a_t)$ represents the state-action value function, which indicates the expected cumulative reward after taking action $a_t$. $\mathcal{H}(\pi(\cdot\mid s_t))$ is the policy entropy that encourages exploration. $\alpha$ is the entropy regularization coefficient, which balances exploration with reward maximization. In SAC, the Q function is updated by minimizing the following mean squared error objective:

\begin{algorithm}[t]
    \caption{Unsupervised pretraining}
    \label{algorithm1}
    \textbf{Input}: signal dataset $\mathbf{x}$, batch size $B$, CL training epochs $E_{cl}$, constant $\tau$, encoder $f_\theta$, $f_\xi$, $f_\phi$, projection head $g(\bullet)$, RL training epochs $E_{rl}$, agent $\pi$, state dimension $s_t$, action dimension $a_t$, reward function $r_t$.\\
    \textbf{Output}: encoder weight $f_\theta$, $f_\xi$, $f_\phi$. \\
    \begin{algorithmic}[1] 
    \STATE Initialize $f_\theta$, $f_\xi$, $f_\phi$, $g(\bullet)$. Initialize agent $\pi$. \\
    \STATE $r_t=0$, $acc_{best}=0$. \\
    \FOR{$\text{t} = 1$ to $E_{rl}$}
        \STATE \textbf{if} $\text{t} = 1$ \textbf{then}
        \STATE \quad Set all action parameters to 0.
        \STATE \textbf{else}
        \STATE \quad Compute $a_t$ through~\cref{eq21}.
        \STATE \quad load agent $\pi$ weight.
        \FOR{$\text{iteration} = 1$ to $E_{cl}$}
            \STATE \quad $\mathbf{z}_i^{T}=g_\theta(f_\theta(\mathbf{x}_i^T))$ and $\tilde{\mathbf{z}}_i^{T}=g_\theta(f_\theta(\tilde{\mathbf{x}}_i^T))$. 
            \STATE \quad $\mathbf{z}_i^{F}=g_\xi(f_\xi(\mathbf{x}_i^F))$ and $\tilde{\mathbf{z}}_i^{F}=g_\xi(f_\xi(\tilde{\mathbf{x}}_i^F))$. 
            \STATE \quad $\mathbf{z}_i^{C}=g_\phi(f_\phi(\mathbf{x}_i^C))$ and $\tilde{\mathbf{z}}_i^{C}=g_\phi(f_\phi(\tilde{\mathbf{x}}_i^C))$.
            \STATE \quad Compute loss through~\cref{eq14,eq15,eq16,eq17,eq18,eq19}.
            \STATE \quad Update $f_\theta$, $f_\xi$, and $f_\phi$ to minimize loss.
        \ENDFOR
        \STATE \quad Compute $acc_t$ through K-means.
        \STATE \quad Compute $r_t$ through~\cref{eq22}.
        \STATE \quad Compute $s_t$ through~\cref{eq20}.
        \STATE \quad training agent $\pi$ through~\cref{eq25,eq26,eq27,eq28}.
        \STATE \quad Update and save agent $\pi$.
        \STATE \textbf{if} $acc_t > acc_{best}$ \textbf{then}
        \STATE \quad $acc_{best} = acc_t$.
        \STATE \quad save encoder weight $f_\theta$, $f_\xi$, $f_\phi$.
    \ENDFOR
    \end{algorithmic}
\end{algorithm}

\begin{equation}
\label{eq26}
J(Q)=\mathbb{E}_{(s_t,a_t)\sim\mathcal{D}}\left[\left(Q(s_t,a_t)-\hat{Q}(s_t,a_t)\right)^2\right],
\end{equation}
where $\hat{Q}(s_t,a_t)$ is the target Q value, calculated using the following formula:
\begin{multline}
\label{eq27}
\hat{Q}(s_t,a_t)=r_t+\gamma\mathbb{E}_{a_{t+1}\sim\pi}\left[Q(s_{t+1},a_{t+1})- \right. \\
\left. \alpha\log\pi(a_{t+1}\mid s_{t+1})\right].
\end{multline}
Finally, the update objective for the policy $\pi$ is to maximize the following expectation:
\begin{equation}
\label{eq28}
J(\pi)=\mathbb{E}_{s_t\sim\mathcal{D},a_t\sim\pi}\left[\alpha\log\pi(a_t\mid s_t)-Q(s_t,a_t)\right].
\end{equation}
By performing gradient optimization updates on the parameterized policy, the optimal policy is obtained.

The pretraining process of the MCLRL framework is described in~\cref{algorithm1}. First, the signal data is processed and divided into three representation domains to be input into the pretraining framework. Three encoders are prepared to extract features from the signals in these domains, along with a projection head and a reinforcement learning SAC agent. In the initial step, all model parameters are initialized, and certain variables are set. During the reinforcement learning loop, if it is the first iteration, the actions are set to zero to avoid manual interference with the agent's decision-making. In subsequent iterations, the actions are determined by the agent, and its weights are loaded accordingly. The contrastive learning training loop is then conducted. Upon completion, K-means is used for accuracy evaluation, and the corresponding reward and state are returned to the agent. Finally, the agent is trained, and the reinforcement training loop is repeated. Only the model weights of the three encoders that achieve optimal results during evaluation are retained.
\subsection{Fine-tuning and Testing}
\begin{figure}[t]
\centering
\includegraphics[width=3.3in, clip, trim=0.3in 0.3in 0.3in 0.3in]{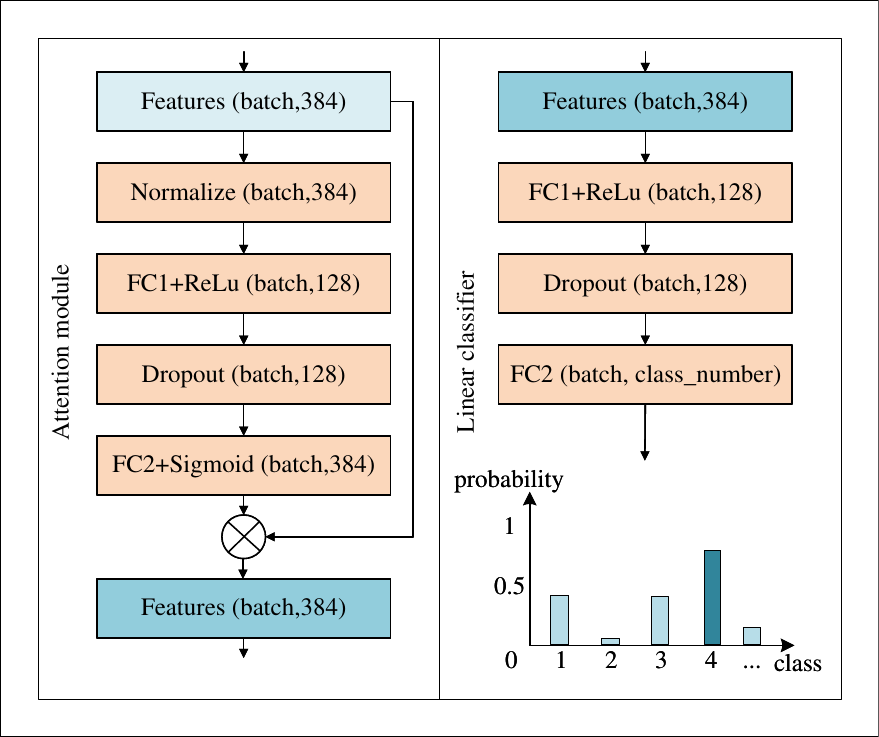}
\caption{The left side of the image depicts the attention module. On the right side is the linear classifier.}
\label{fig_3}
\end{figure}

In the previous unsupervised pretraining phase, the weight parameters of the three encoders are obtained. In the supervised fine-tuning phase, the first step is to freeze the weights of the three encoders. Since fine-tuning the attention module and classifier is done with only a minimal number of samples and a very short training cycle, too many model weights could lead to overfitting. Additionally, the output features of the last layer of the three models should be set to 128-dimensional vectors. This ensures balanced learning across the models without bias toward any one model. During the fine-tuning phase, the features from the three representation domains are directly concatenated instead of being summed or multiplied to prevent the loss of critical information~\cite{zha2024few}. The expression for direct concatenation is:
\begin{equation}
\label{eq29}
\mathbf{C}=\mathrm{concat}(f_\theta(\mathbf{x}_i^T),f_\xi(\mathbf{x}_i^F),f_\phi(\mathbf{x}_i^C)),
\end{equation}
where $\mathbf{C}\in\mathbb{R}^{B\times 384}$ represents the feature matrix. The concatenated features, as shown in Fig.~\ref{fig_3}, which are directly fed into the attention module. This module uses a small number of linear and activation layers to assign weights to the features from each representation domain, effectively preventing overfitting. Subsequently, the weighted features are passed to the linear classifier, which classifies the features into the desired categories, ultimately yielding the probability for each category.

In the final downstream task classification phase, we should freeze all the parameters of the models and modules, performing only evaluation operations. The results from the linear classifier are compared with the ground truth labels, and the cross-entropy loss function is used. This function is widely employed in classification problems, especially in multi-class classification tasks, as it effectively measures the difference between the model's predictions and the actual labels. The mathematical expression of the cross-entropy loss function is as follows:
\begin{equation}
\label{eq30}
L_{cross}=-\sum_iy_i\mathrm{log}(\hat{y_i}),
\end{equation}
where $y_i$ represents the true label, and $\hat{y_i}$ denotes the predicted probability by the model. This loss function improves the model’s classification performance by minimizing the discrepancy between the predicted probability and the actual label.

\section{Experiments}
\label{sec:Experiments}
\subsection{Dataset Introduction}
To evaluate the effectiveness of our method, we conduct experiments on signal modulation classification datasets, including RML2016.10a~\cite{o2016radio}, Sig2019-12~\cite{chen2021signet}, RML2016.10a (high), and Sig2019-12 (high).

\begin{figure*}[htbp]
    \centering
    \begin{minipage}[b]{0.48\textwidth}  
        \centering
        \includegraphics[clip, trim=0.1in 0.1in 0.1in 0.1in, width=\textwidth]{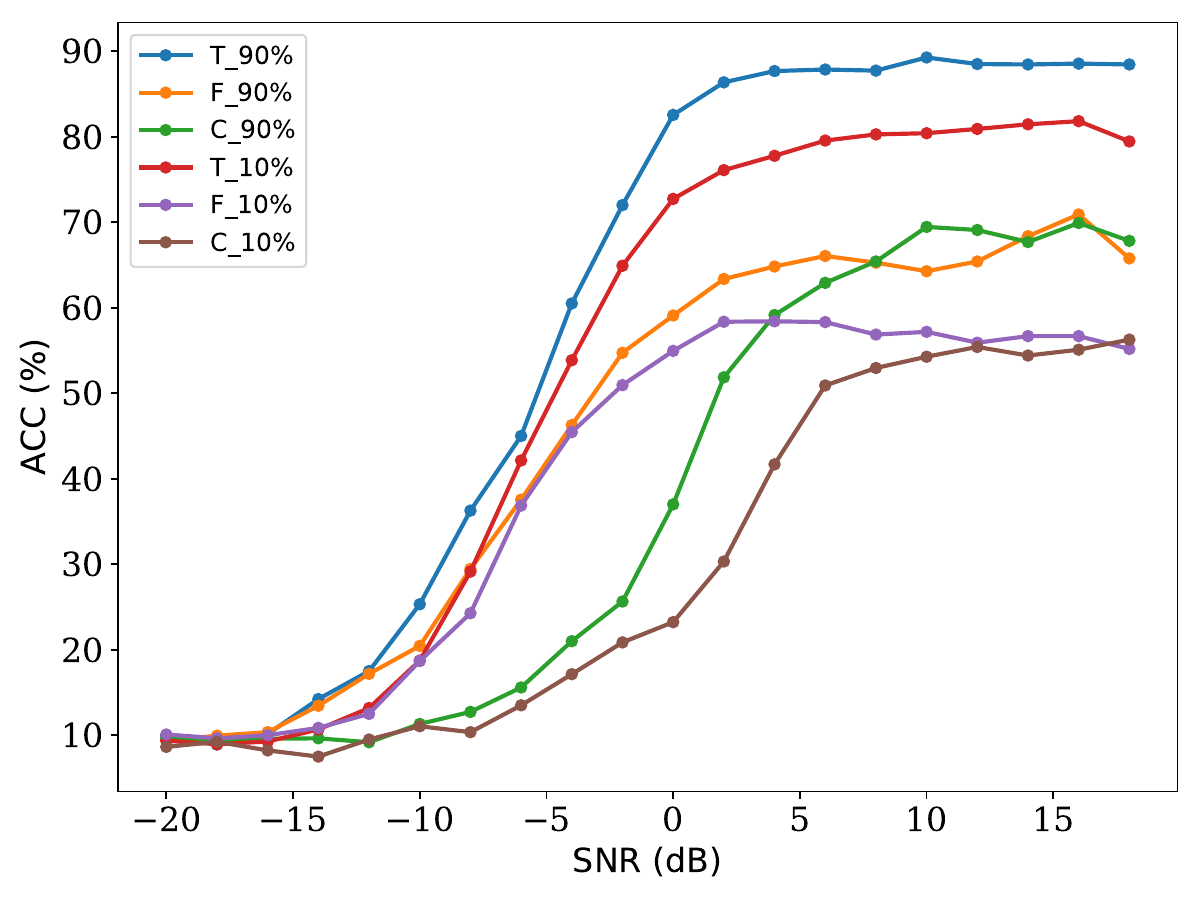}
        \subfloat{(a)}
    \end{minipage}
    \hspace{0.00\textwidth}  
    \begin{minipage}[b]{0.48\textwidth}
        \centering
        \includegraphics[clip, trim=0.1in 0.1in 0.1in 0.1in, width=\textwidth]{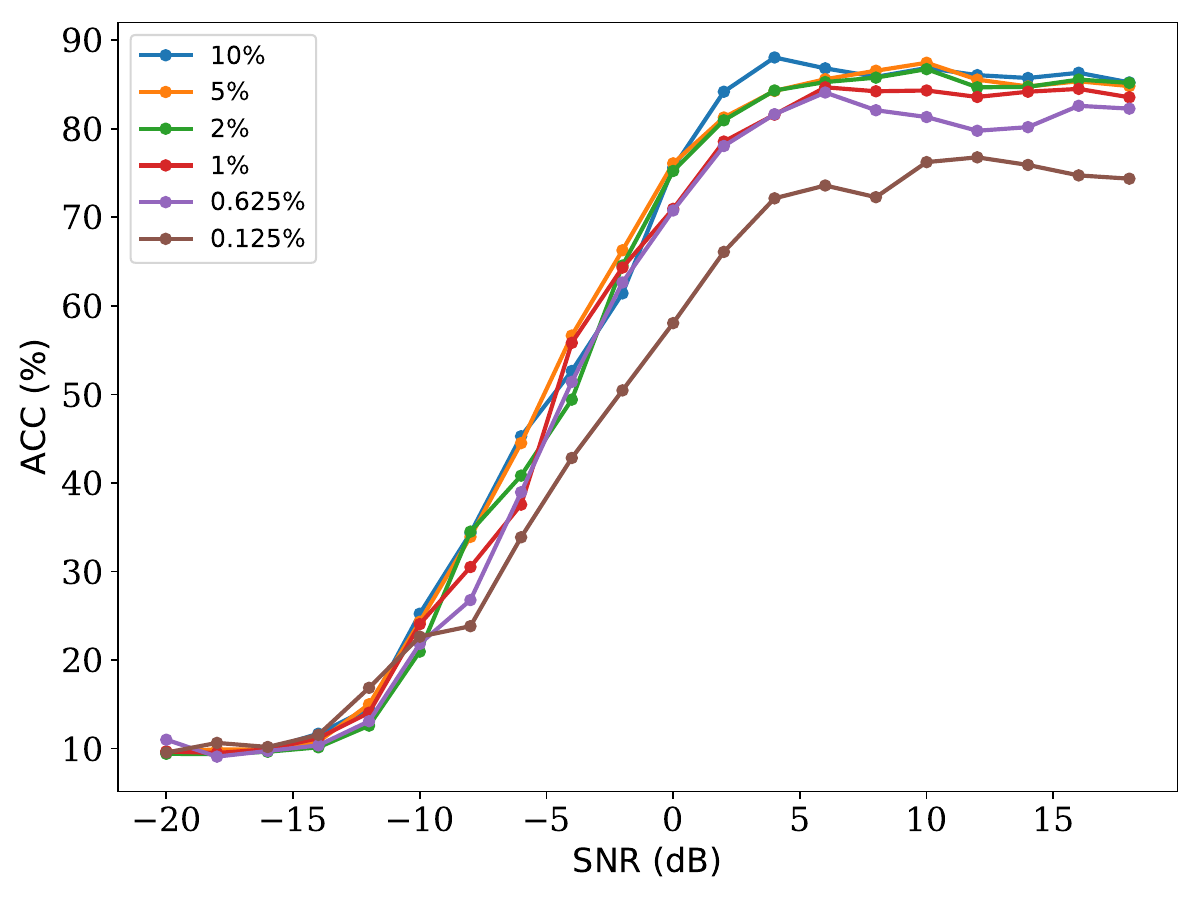}
        \subfloat{(b)}   
    \end{minipage}
    \hspace{0.00\textwidth}  
    \begin{minipage}[b]{0.48\textwidth}
        \centering
        \includegraphics[clip, trim=0.1in 0.1in 0.1in 0.1in, width=\textwidth]{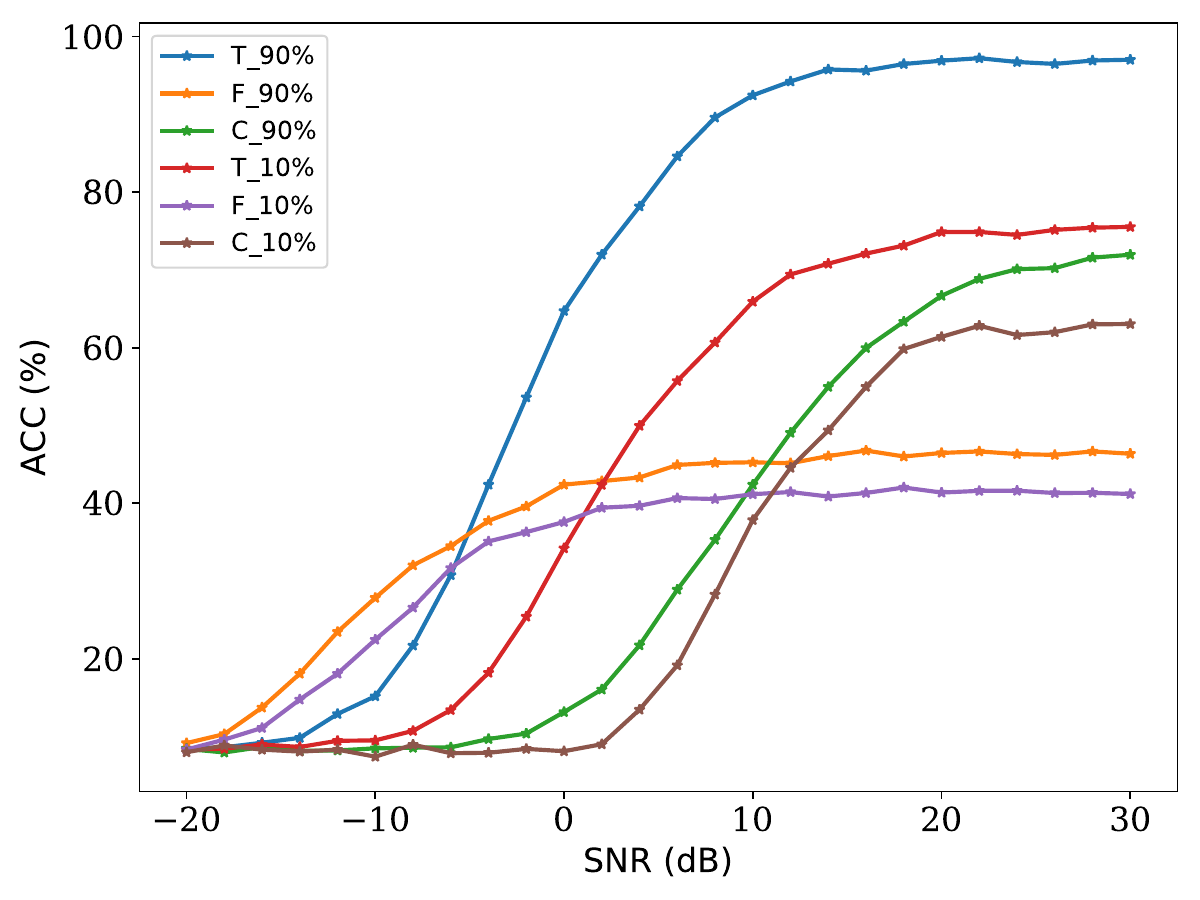}
        \subfloat{(c)}   
    \end{minipage}
    \hspace{0.00\textwidth}  
    \begin{minipage}[b]{0.48\textwidth}
        \centering
        \includegraphics[clip, trim=0.1in 0.1in 0.1in 0.1in, width=\textwidth]{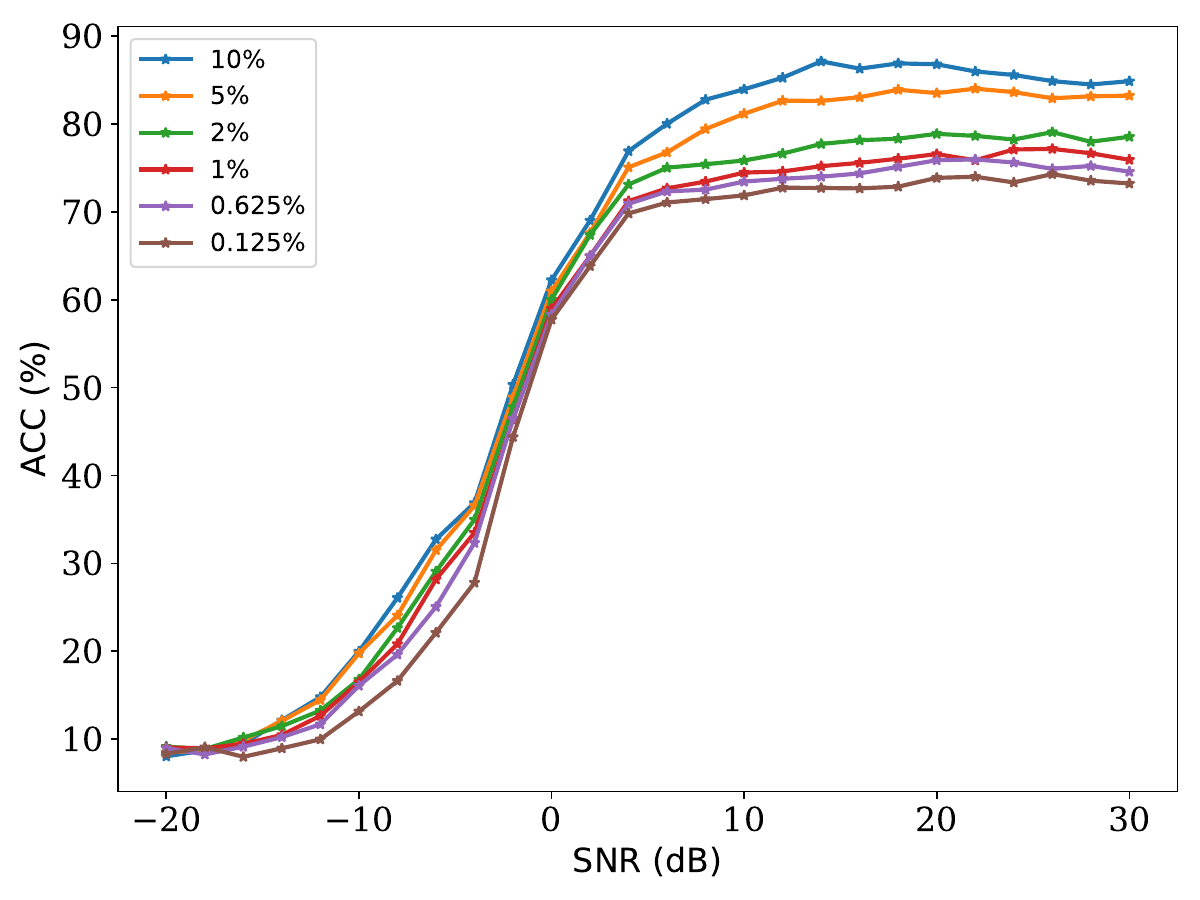}
        \subfloat{(d)}   
    \end{minipage}
    \caption{(a) and (b) represent the RML2016.10a dataset, indicated by the '-o-' symbol, while (c) and (d) represent the Sig2019-12 dataset, denoted by the '-*-' symbol. Taking RML2016.10a as an example, (a) shows the encoder trained in a supervised manner, where 'T\_90\%' indicates that the time domain encoder uses 90\% of the training set samples. 'F' denotes the frequency domain encoder, while 'C' denotes the constellation diagram encoder. (b) shows the MCLRL framework in an unsupervised FSL setting, where '0.125\%' represents the 1-shot sample size used for fine-tuning.}
    \label{fig_4}
\end{figure*}

\textbf{RML2016.10a} is a comprehensive public radio signal dataset generated using GNU Radio~\cite{blossom2004gnu}, containing $220,000$ signals.  It includes 11 modulation types, comprising eight digital variations—BPSK, QPSK, 8PSK, 16QAM, 64QAM, BFSK, CPFSK, and PAM4—and three analog ones—WB-FM, AM-SSB, and AM-DSB.  Each modulation type is associated with 20 distinct signal-to-noise ratios (SNR), ranging from $-20$dB to $18$dB in $2$dB intervals, with each SNR level containing $1,000$ samples. Each signal has a length of 128, and the dataset is divided into a training set and a test set in an 8:2 ratio. The SNR value represents signal quality, with higher SNR indicating less noise interference.

\textbf{RML2016.10a (high)} is similar to the RML2016.10a dataset in all other aspects. The SNR ranges from 10 dB to 18 dB, in intervals of 2 dB. The training and test sets contain 44,000 and 11,000 samples, respectively.

\textbf{Sig2019-12} is a private collection simulated by Chen et al.~\cite{chen2021signet} and contains $468,000$ signals across 12 modulation types: BPSK, QPSK, 8PSK, OQPSK, 2FSK, 4FSK, 8FSK, 16QAM, 32QAM, 64QAM, 4PAM, and 8PAM. The SNR ranges from $-20$dB to $30$dB in $2$dB intervals, with each signal consisting of 512 sampling points derived from 64 symbols oversampled at a rate of 8. The dataset factors in various complex communication system aspects, such as carrier phase, pulse shaping, frequency offsets, and noise. It is partitioned into a training set and a test set in a 2:1 ratio. Each modulation type includes $1,500$ samples per SNR.

\textbf{Sig2019-12 (high)} is similar to the Sig2019-12 dataset in all other aspects. The SNR for each modulation type is uniformly distributed from 10 dB to 30 dB, with a 2 dB interval. The training set consists of 132,000 samples, while the test set contains 66,000 samples.

\subsection{Experimental Setup}
\textit{1) Model and Dataset:} The proposed MCLRL framework is implemented in PyTorch and trained on a Tesla V100 GPU. To validate the effectiveness of the MCLRL framework, we use ResNet1D~\cite{chen2021signet}, a typical network model designed for AMR tasks, as the encoder for the time and frequency domains, and ResNet18~\cite{he2016deep} as the encoder for the constellation diagram. These encoders are used to extract features from each respective representation domain. The output features of all encoders are reduced to 128 dimensions. The reinforcement learning model used is the SAC agent. For data processing, we use 90 percent of the training set from the full SNR and full category dataset, without labels, as the $base$ set, while the remaining 10 percent is kept as the labeled $support$ set, with the test set serving as the $query$ set.

\textit{2) Training Setup:} The Adam optimizer is used, with the learning rate set to 0.001 during both the pretraining and fine-tuning stages for training our framework. In the pretraining stage, the contrastive learning cycle is set to 30 epochs, and the reinforcement learning cycle is set to 10 epochs. In the downstream fine-tuning stage, only 10 epochs are used to achieve fast fine-tuning. Based on the setup, the batch sizes of the input samples vary for the $base$ set, $support$ set, and $query$ set. The temperature coefficient $\tau=0.05$ for the loss function and the hyperparameter $\lambda=0.8$ are also considered in the configuration.

Other details of the experimental setup are elaborated in the following subsections. In Subsection~\ref{sec:basic}, we compare the unsupervised framework of MCLRL with the supervised model and provide a detailed introduction to the setup of the FSL task. In Subsection~\ref{sec:Contrast}, we compare MCLRL with some commonly used CL and FSL methods. In Subsection~\ref{sec:Ablation}, we conduct ablation experiments on some modules and parameters. In Subsection~\ref{sec:Visualization}, we visualize the features extracted by the MCLRL framework to validate their effectiveness.

\newcommand{\Xhline}[1]{\noalign{\hrule height #1}}
\renewcommand{\arraystretch}{1.5}
\begin{table*}[htbp]  
    \centering
    \caption{\MakeUppercase{The comparative experimental results on two different datasets are shown, with performance validation of various methods using small-batch fine-tuning in downstream tasks, which is commonly used in FSL tasks. The bolded values indicate performance superior to other methods.}}  
    \label{table_1}
    \resizebox{\textwidth}{!}{  
        \begin{tabular}{c|ccccccc|ccccccc}
\Xhline{1.5pt}
\multirow{2}{*}{Method} & \multicolumn{6}{c}{RML2016.10a}                                                                         &                & \multicolumn{6}{c}{Sig2019-12}                                                                          &                \\ \cline{2-15} 
                        & 1-shot (0.125\%) & 5-shot (0.625\%) & 8-shot (1\%)   & 16-shot (2\%)  & 40-shot (5\%)  & 80-shot (10\%) & Avg ACC       & 1-shot (0.125\%) & 5-shot (0.625\%) & 8-shot (1\%)   & 16-shot (2\%)  & 40-shot (5\%)  & 80-shot (10\%) & Avg ACC       \\ \Xhline{1.5pt}
SimCLR (PMLR, 2020)                  & 38.32            & 41.98            & 43.15          & 44.68          & 45.67          & 45.73          & 43.26          & 29.81            & 37.29            & 40.14          & 42.20          & 44.11          & 45.24          & 39.80          \\
MoCoV2 (CVPR, 2020)                 & 39.11            & 43.16            & 45.84          & 47.14          & 49.45          & 51.26          & 45.99          & 31.13            & 32.49            & 36.94          & 39.15          & 45.13          & 46.30          & 38.52          \\
BYOL (NeurIPS, 2020)                    & 35.26            & 42.85            & 43.86          & 44.45          & 44.91          & 45.22          & 42.76          & 22.62            & 28.71            & 31.45          & 33.58          & 33.17          & 37.11          & 31.11          \\
TS-TCC (ML, 2021)                  & 35.78            & 40.34            & 47.96          & 48.46          & 50.19          & 52.11          & 45.81          & 21.76            & 23.64            & 27.40          & 29.98          & 31.60          & 33.57          & 27.99          \\
TF-C (NeurIPS, 2022)                   & 24.51            & 41.21            & 42.11          & 47.98          & 48.78          & 51.72          & 42.72          & 16.57            & 18.10            & 21.63          & 25.38          & 33.94          & 38.43          & 25.68          \\
TS2Vec (AAAI, 2022)                 & 29.40            & 37.91            & 40.54          & 45.63          & 47.32          & 49.71          & 41.75          & 22.29            & 25.47            & 30.22          & 35.40          & 37.22          & 40.05          & 31.78          \\
DFR (TNNLS, 2023)                    & 42.41            & 44.27            & 48.55          & 51.02          & 54.27          & 54.37          & 49.15          & 43.55            & 44.53            & 48.01          & 51.79          & 56.03          & 56.42          & 50.06          \\
FFFNet (TVT, 2024)                  & 44.94            & 47.27            & 52.65          & 52.95          & 53.97          & 54.74          & 51.09          & 45.23            & 47.87            & 51.15          & 53.17          & 55.51          & 56.07          & 51.50          \\
ScSP (IoT, 2024)                   & 46.13            & 50.92            & 54.01          & \textbf{55.73} & 56.06          & 56.09          & 53.16          & 49.90            & 51.61            & 52.24          & \textbf{55.60} & 56.39          & 58.66          & 54.07          \\
Ours                    & \textbf{47.62}   & \textbf{52.88}   & \textbf{54.33} & 54.99          & \textbf{56.13} & \textbf{56.25} & \textbf{53.70} & \textbf{50.29}   & \textbf{51.91}   & \textbf{52.79} & 54.35          & \textbf{57.23} & \textbf{58.93} & \textbf{54.25} \\ \hline
Better than the second                       & 1.49             & 1.96             & 0.32           & -0.74          & 0.07           & 0.16           & 0.54           & 0.39             & 0.30             & 0.55           & -1.25          & 0.84           & 0.27           & 0.18           \\ \Xhline{1.5pt}
\multirow{2}{*}{}       & \multicolumn{6}{c}{RML2016.10a (high)}                                                                  &                & \multicolumn{6}{c}{Sig2019-12 (high)}                                                                   &                \\ \cline{2-15} 
                        & 1-shot (0.125\%) & 5-shot (0.625\%) & 8-shot (1\%)   & 16-shot (2\%)  & 40-shot (5\%)  & 80-shot (10\%) & Avg ACC       & 1-shot (0.125\%) & 5-shot (0.625\%) & 8-shot (1\%)   & 16-shot (2\%)  & 40-shot (5\%)  & 80-shot (10\%) & Avg ACC       \\ \Xhline{1.5pt}
SimCLR (PMLR, 2020)                  & 53.18            & 57.73            & 59.82          & 61.82          & 64.36          & 65.73          & 60.44          & 56.82            & 66.87            & 69.20          & 71.15          & 75.63          & 77.27          & 69.49          \\
MoCoV2 (CVPR, 2020)                  & 64.86            & 67.82            & 71.44          & 73.60          & 78.77          & 79.59          & 72.68          & 64.68            & 65.97            & 69.58          & 73.33          & 78.05          & 79.13          & 71.79          \\
BYOL (NeurIPS, 2020)                    & 55.45            & 59.41            & 64.19          & 65.64          & 67.91          & 68.32          & 63.49          & 40.25            & 46.45            & 56.05          & 58.88          & 60.90          & 65.90          & 54.74          \\
TS-TCC (ML, 2021)                  & 60.09            & 63.05            & 69.05          & 71.36          & 72.50          & 76.41          & 68.74          & 41.13            & 45.22            & 53.58          & 55.88          & 61.52          & 63.80          & 53.52          \\
TF-C (NeurIPS, 2022)                    & 63.14            & 65.64            & 67.32          & 69.14          & 71.64          & 73.64          & 68.42          & 29.77            & 32.38            & 52.87          & 56.65          & 61.05          & 64.98          & 49.62          \\
TS2Vec (AAAI, 2022)                  & 57.59            & 60.09            & 66.95          & 68.41          & 70.09          & 71.59          & 65.79          & 44.56            & 47.62            & 53.17          & 60.23          & 64.63          & 65.58          & 55.97          \\
DFR (TNNLS, 2023)                     & 59.23            & 65.32            & 71.18          & 76.95          & 77.27          & 80.18          & 71.69          & 60.05            & 64.23            & 67.93          & 71.98          & 77.15          & 80.50          & 70.31          \\
FFFNet (TVT, 2024)                  & 64.27            & 69.09            & 74.55          & 79.68          & 80.27          & 81.86          & 74.95          & 67.07            & 70.90            & 73.67          & 76.08          & 79.93          & 82.88          & 75.09          \\
ScSP (IoT, 2024)                    & 71.91            & 75.77            & 79.59          & 82.18          & 82.55          & 82.86          & 79.14          & 68.33            & 70.57            & 73.10          & 77.55          & 80.23          & 83.87          & 75.61          \\
Ours                    & \textbf{75.60}   & \textbf{81.23}   & \textbf{84.03} & \textbf{85.37} & \textbf{85.59} & \textbf{86.04} & \textbf{82.98} & \textbf{73.20}   & \textbf{74.81}   & \textbf{75.93} & \textbf{78.01} & \textbf{83.08} & \textbf{85.65} & \textbf{78.45} \\ \hline
Better than the second                       & 3.69             & 5.46             & 4.44           & 3.19           & 3.04           & 3.18           & 3.83           & 4.87             & 4.24             & 2.83           & 0.46           & 2.85           & 1.78           & 2.84           \\ \Xhline{1.5pt}
\end{tabular}
    }
\end{table*}

\subsection{Basic Experiment}
\label{sec:basic}
In this experiment, we compare the performance of supervised training with three different domain encoders and unsupervised training using varying sample sizes. 'T\_90\%' indicates that the time domain encoder is trained with 90\% of the training set samples in a supervised manner, and its performance is tested on the test set, with the same procedure applied to the other encoders. In unsupervised training, the $base$ set is used for unsupervised training. For the FSL task, we set up six groups with different sample quantities: 10\% (11-way-80shot), 5\% (11-way-40shot), 2\% (11-way-16shot), 1\% (11-way-8shot), 0.625\% (11-way-5shot), and 0.125\% (11-way-1shot), to validate performance under varying sample conditions. The final performance validation is conducted on two completely different datasets, as shown in Fig.~\ref{fig_4}.

As shown in Figure (a) and (b), in the RML2016.10a dataset, models trained using traditional time domain encoders and a large amount of labeled data achieve an accuracy of 90\% on high SNR data, while the frequency domain encoder and constellation diagram encoder only reach around 70\% accuracy. This indicates that frequency domain features and image-based features have limitations in modulation signal classification. Once the amount of supervised training data is reduced to 10\%, the accuracy drops by 10\%-15\%. In contrast, the encoders trained in an unsupervised manner within the MCLRL framework maintain an accuracy between 80\%-90\% on high SNR data, even with a very small number of samples for fine-tuning. Specifically, the one-shot case, due to the extremely small sample size, suffers from poor generalization performance, but still outperforms the supervised training of the frequency domain and constellation diagram models. As shown in Figure (c) and (d), the results in the Sig2019-12 dataset are similar to those in the RML2016.10a dataset. In this dataset, the frequency domain encoder performs better than the other methods under low SNR conditions, but consistently underperforms at high SNR. Ultimately, these results demonstrate that the MCLRL framework can still maintain high classification performance even with a minimal amount of labeled samples.

\begin{table}[]
\centering
\caption{\MakeUppercase{The ablation experimental results of the MCLRL framework's loss functions are shown, with 5-shot sample size used for downstream fine-tuning on two datasets.}}  
\label{table_2}
\begin{tabular}{cccc|c|c}
\Xhline{1.5pt}
\multicolumn{4}{c|}{Loss Type}        & RML2016.10a    & Sig2019-12     \\ \hline
Loss\_1 & Loss\_2 & Loss\_3 & Loss\_4 & ACC(\%)        & ACC(\%)        \\ \Xhline{1.5pt}
\checkmark       &         &         &         & 44.60          & 46.52          \\
\checkmark       & \checkmark       &         &         & 52.21          & 50.47          \\
\checkmark       &         & \checkmark       &         & 50.83          & 48.18          \\
\checkmark       & \checkmark       & \checkmark       &         & \textbf{52.88} & \textbf{51.91} \\
\checkmark       & \checkmark       & \checkmark       & \checkmark       & 44.78          & 44.68          \\ \Xhline{1.5pt}
\end{tabular}
\end{table}

\begin{table}[htbp]
\centering
\caption{\MakeUppercase{The ablation experimental results of the data augmentation types in the MCLRL framework are presented.}}  
\label{table_3}
\resizebox{\columnwidth}{!}{  
    \begin{tabular}{ccccc|c|c}
        \Xhline{1.5pt}
        \multicolumn{5}{c|}{Data Augmentation}                                    & RML2016.10a    & Sig2019-12     \\ \hline
        noise addition & time shifting & scaling & random dropout & interpolation & ACC(\%)        & ACC(\%)        \\ \Xhline{1.5pt}
        \checkmark              &               &         &                &               & 51.01          & 49.69          \\
        \checkmark              & \checkmark             &         &                &               & 51.33          & 51.02          \\
        \checkmark              & \checkmark             & \checkmark       &                &               & 51.17          & 51.72          \\
        \checkmark              & \checkmark             & \checkmark       & \checkmark              &               & 52.67          & 51.69          \\
        \checkmark              & \checkmark             & \checkmark       & \checkmark              & \checkmark             & \textbf{52.88} & \textbf{51.91} \\ \Xhline{1.5pt}
    \end{tabular}
}
\end{table}

\begin{figure}[t]
\centering
\includegraphics[width=3.3in, clip, trim=0.1in 0.1in 0.1in 0.1in]{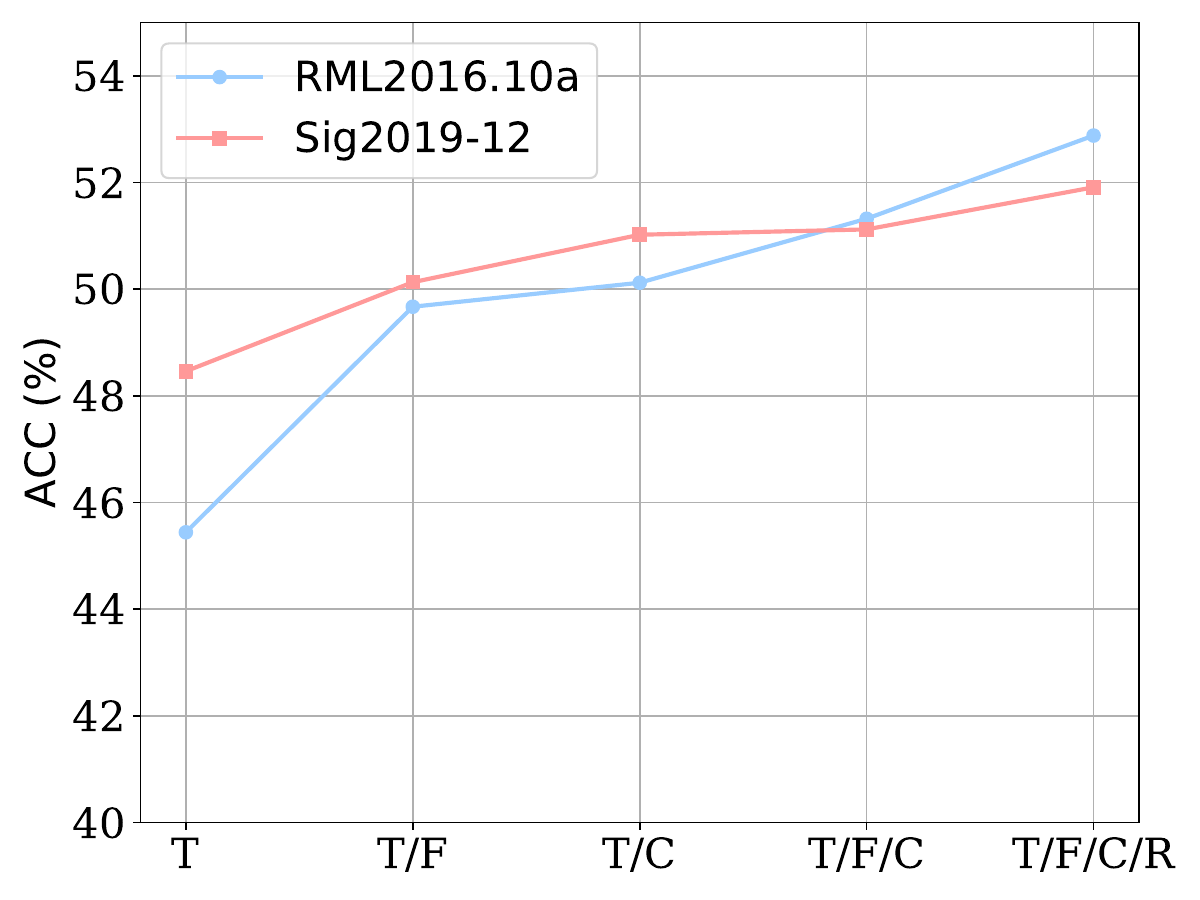}
\caption{The experimental results on the effectiveness of multi-representation domains and reinforcement learning in the MCLRL framework are shown. 'T' represents the time domain, 'F' represents the frequency domain, and 'C' represents the constellation diagram domain. Whether reinforcement learning is used is indicated by 'R'.}
\label{fig_5}
\end{figure}

\begin{figure}[t]
\centering
\includegraphics[width=3.3in, clip, trim=0.1in 0.1in 0.1in 0.1in]{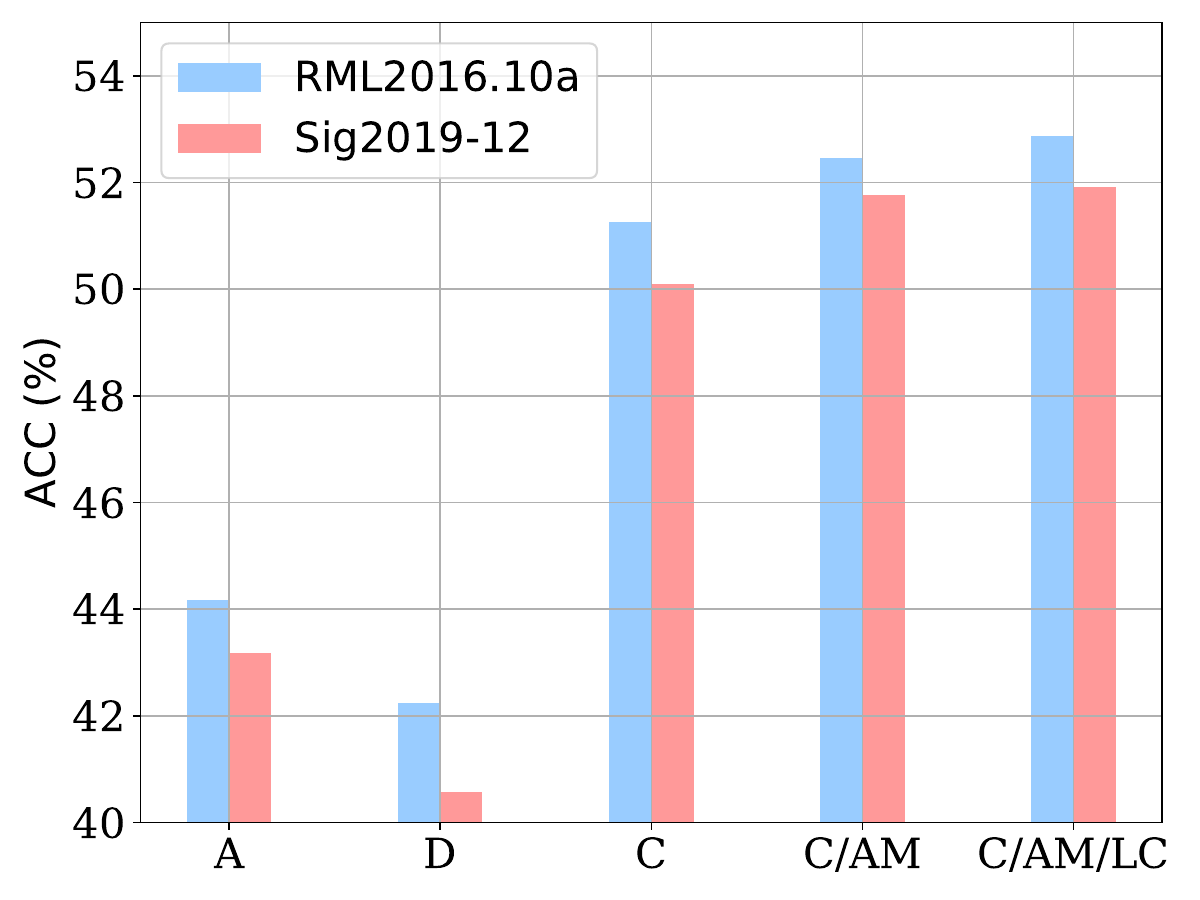}
\caption{The experimental results on the performance of feature fusion methods, attention modules, and linear classifiers during the downstream task phase are shown. The feature fusion methods are: 'A' for addition, 'D' for dot product, and 'C' for direct concatenation. The use of the attention module and linear classifier is denoted by 'AM' and 'LC',  respectively.}
\label{fig_6}
\end{figure}

\subsection{Contrast Experiment}
\label{sec:Contrast}
In the comparison experiments, we selected three contrastive learning algorithms from the image domain for modulation signal classification, namely SimCLR~\cite{chen2020simple}, MoCoV2~\cite{chen2020improved}, and BYOL~\cite{grill2020bootstrap}. Additionally, for the signal domain, we chose three time-series contrastive learning algorithms, specifically TS-TCC~\cite{eldele2021time}, TF-C~\cite{zhang2022self}, and TS2Vec~\cite{yue2022ts2vec}. Finally, for the FSL task, we applied several FSL methods to modulation signal for comparison, including FFFNet~\cite{zha2024few}, DFR~\cite{cheng2023disentangled}, and ScSP~\cite{su2024semantic}.

To facilitate performance evaluation and ensure data comparability, we uniformly employed ResNet1D as the feature extraction model across all methods. In the pretraining phase, we consistently used the $base$ set, while the $support$ set is used for testing, evaluating performance under different sample size conditions. Due to significant differences in training methodologies, we did not apply a unified approach to pretraining cycles and batch sizes; instead, we selected the optimal hyperparameters for each experiment.

The experimental results are presented in Table~\ref{table_1}. The earliest contrastive learning methods for image domains, such as SimCLR, MoCoV2, and BYOL, did not exhibit outstanding performance in the modulation signal FSL task. Whether on the RML2016.10a or Sig2019-12 dataset, these methods required a larger number of samples in the downstream task phase to achieve good performance. On the other hand, time-series methods like TS-TCC, TF-C, and TS2Vec performed poorly in the FSL task. For methods specifically designed for the FSL task, such as FFFNet, DFR, and ScSP, there are some limitations. On one hand, these methods did not fully exploit all the features of modulation signal during feature extraction. On the other hand, in the downstream fine-tuning phase, particularly in the extreme cases of 1-shot and 5-shot, their generalization performance is poor due to overly simplistic features. However, the ScSP method demonstrated a leading advantage at 16-shot. In contrast, the MCLRL framework showed clear advantages in both 1-shot and 5-shot scenarios. As the number of fine-tuning samples increased, the accuracy of the other methods gradually approached MCLRL's performance.

The MCLRL framework shows a more significant advantage in the high SNR datasets, such as RML2016.10a (high) and Sig2019-12 (high). This is because MCLRL does not specifically handle low SNR data, resulting in lower accuracy at low SNR levels. However, at high SNR levels, where the signal noise is smaller, the features are more distinguishable. Therefore, after fully leveraging the features from the three representation domains, the framework performs better in classification. The average accuracy improves by 3\% to 15\%.

\begin{table*}[htbp]  
    \centering
    \caption{\MakeUppercase{The ablation experimental results of the time domain encoder on the RML2016.10a dataset are shown, with the other two encoders remaining unchanged. The metric is the accuracy (\%) for each SNR.}}  
    \label{table_4}
    \resizebox{\textwidth}{!}{  
        \begin{tabular}{c|cccccccccccccccccccc|c}
        \Xhline{1.5pt}
        \multirow{2}{*}{T\_Encode} & \multicolumn{20}{c|}{SNR (dB)}                                                                                                                                & \multirow{2}{*}{Avg ACC} \\ \cline{2-21}
                                   & -20  & -18   & -16   & -14   & -12   & -10   & -8    & -6    & -4    & -2    & 0     & 2     & 4     & 6     & 8     & 10    & 12    & 14    & 16    & 18    &                           \\ \Xhline{1.5pt}
        ResNet1D                   & 11   & 9.09  & 9.73  & 10.36 & 13.09 & 21.82 & 26.77 & 38.95 & 51.41 & 62.64 & 70.77 & 78.05 & 81.64 & 84.09 & 82.09 & 81.32 & 79.77 & 80.18 & 82.59 & 82.27 & 52.88                     \\
        CNN2D                      & 9.95 & 8.91  & 9.14  & 11.23 & 12.5  & 20.32 & 26.27 & 35    & 52.5  & 60    & 72    & 78.55 & 79.36 & 81.68 & 80.86 & 79.73 & 80.27 & 81.5  & 82.45 & 82.55 & 52.24                     \\
        AlexNet                    & 9.36 & 10.95 & 10.36 & 13.45 & 13.18 & 21.45 & 25.45 & 37.55 & 56.27 & 64.73 & 69.09 & 73.36 & 74.82 & 86.05 & 85.27 & 78.27 & 77.41 & 78.36 & 80.91 & 80.77 & 52.35                     \\
        LeNet                      & 9.64 & 9.41  & 9.59  & 10.64 & 10.18 & 21.32 & 22.73 & 35.59 & 51    & 65.64 & 67    & 71.86 & 79.14 & 82.91 & 82.41 & 79.45 & 79.09 & 82.68 & 79.91 & 82.82 & 51.65                     \\ \Xhline{1.5pt}
        \end{tabular}
    }
\end{table*}

\subsection{Ablation Experiment}
\label{sec:Ablation}
Here, we conduct ablation experiments on the MCLRL framework to explore the effectiveness of its components, including the loss functions, data augmentation methods, reinforcement learning, and the lightweight attention module. All ablation experiments use a 5-shot sample size for validation, with other hyperparameters consistent with those in the base experiments.

\textit{1) \textbf{Loss Experiment}:} The MCLRL framework involves contrastive learning across multiple representation domains, resulting in four distinct types of contrastive losses. 'Loss\_1' represents the contrastive loss between the original and augmented samples within the same representation domain. 'Loss\_2' refers to the contrastive loss between original samples across different representation domains. 'Loss\_3' is the contrastive loss between augmented samples across different representation domains, and 'Loss\_4' is the contrastive loss between original and augmented samples across different representation domains.

The experimental results, shown in Table~\ref{table_2}, indicate that using only 'Loss\_1' leads to accuracies of 44.6\% and 46.52\% on the RML2016.10a and Sig2019-12 datasets, respectively. This suggests that a single intra-domain loss does not effectively link multi-domain features, and contrastive learning within a single domain alone is insufficient. Combining 'Loss\_1' and 'Loss\_2' increases the accuracies to 52.21\% and 50.47\%, respectively. This suggests that the loss function connects original samples across domains, facilitating effective mutual learning within the same feature space. However, combining 'Loss\_1' and 'Loss\_3' leads to a decrease in accuracy, indicating that the relationship between augmented samples across domains is weaker than between the original samples. Combining 'Loss\_1', 'Loss\_2', and 'Loss\_3' further enhances the learning between feature representations in the same space. After adding 'Loss\_4', accuracy decreases significantly, likely due to the large disparity between the original and augmented samples across domains. Forcing these samples to be treated as positive pairs and pulling them closer in feature space may reduce category differences, leading to performance degradation.

\textit{2) \textbf{Data Augmentation Experiment}:} In this experiment, we investigate the impact of the number of data augmentation methods on the performance of the MCLRL framework. We apply five different data augmentation methods across three distinct representation domains. In the constellation diagram domain, we substituted similar operations for those not originally available.

The experimental results are presented in Table~\ref{table_3}. In both the RML2016.10a and Sig2019-12 datasets, accuracy increases gradually with the number of data augmentation methods, but the improvement is not substantial. This suggests that an increase in the variety of data augmentation methods helps with the extraction of hidden features in modulation signals. A larger number of augmentation methods, or specific augmentation techniques, can effectively enhance the ability to mine the hidden features of modulation signals.

\textit{3) \textbf{Multi-Domain and Reinforcement Learning Experiment}:} In this experiment, we explore the effectiveness of multi-representation domains and reinforcement learning in the MCLRL framework. The representation domains in this framework are 'T' for the time domain, 'F' for the frequency domain, and 'C' for the constellation diagram. Reinforcement learning is an external framework used to select data augmentation methods and their levels, denoted by 'R'. We will progressively incorporate these representation domains and the reinforcement learning framework to validate their effectiveness.

The experimental results are presented in Fig.~\ref{fig_5}. In the RML2016.10a and Sig2019-12 datasets, using only time-domain representation features and manually defined levels of data augmentation results in relatively low accuracy. This indicates that relying solely on the time domain does not effectively capture the deep, hidden features of modulation signals. As the frequency domain and constellation diagram are incorporated, the feature representation of the signal becomes richer, leading to a significant improvement in accuracy. Finally, by replacing manually defined hyperparameters with reinforcement learning to select data augmentation methods and levels, deeper differences between categories can be explored. The reinforcement learning agent, by responding to rewards and environmental states associated with different data augmentation methods, can effectively take actions to identify optimal hyperparameters, thereby enhancing both accuracy and efficiency.

\begin{figure*}[htbp]
    \centering
    \begin{minipage}[b]{0.24\textwidth}  
        \centering
        \includegraphics[clip, trim=0.1in 0.1in 0.1in 0.1in, width=\textwidth]{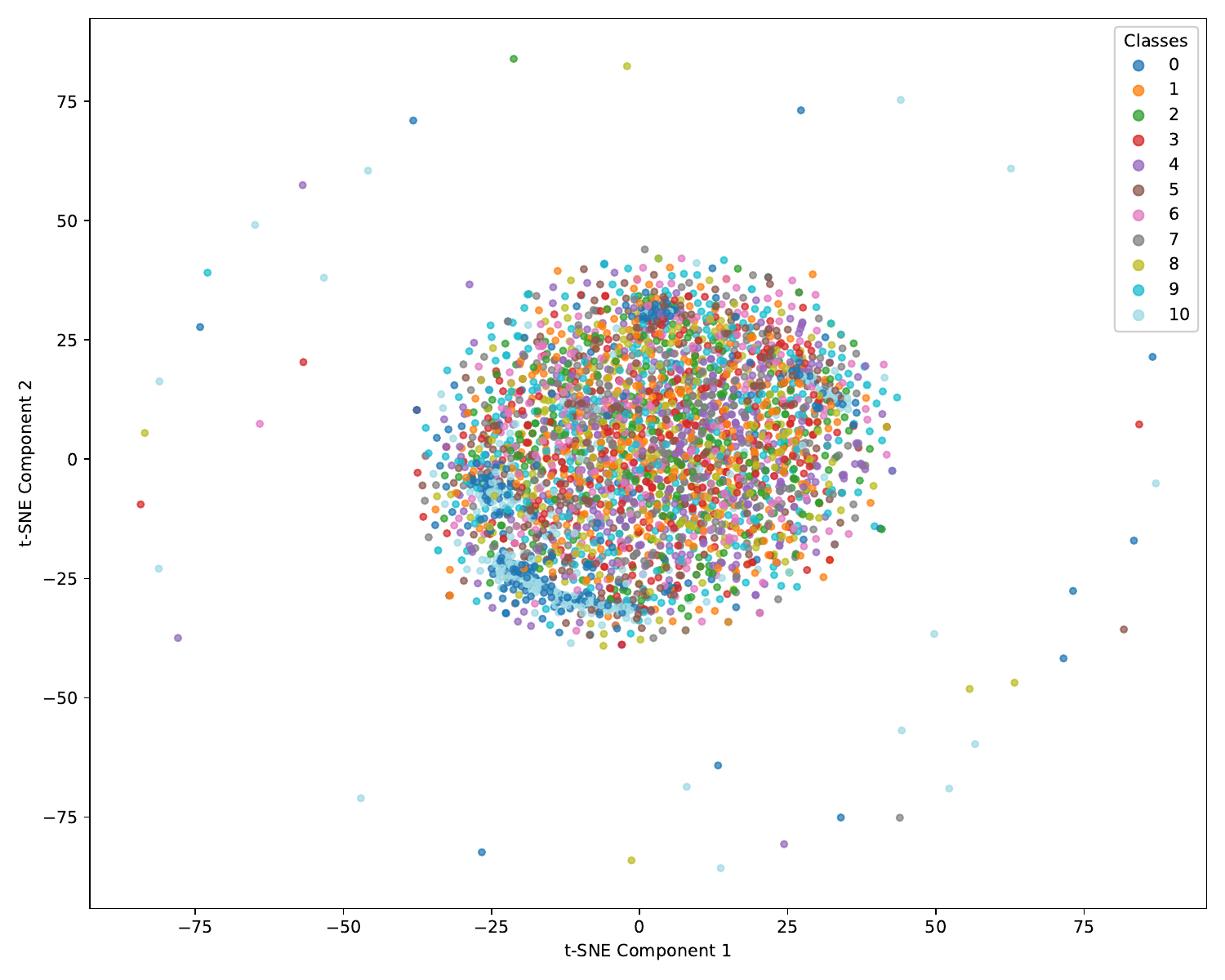}
        \subfloat{(a)}
    \end{minipage}
    \hspace{0.00\textwidth}  
    \begin{minipage}[b]{0.24\textwidth}
        \centering
        \includegraphics[clip, trim=0.1in 0.1in 0.1in 0.1in, width=\textwidth]{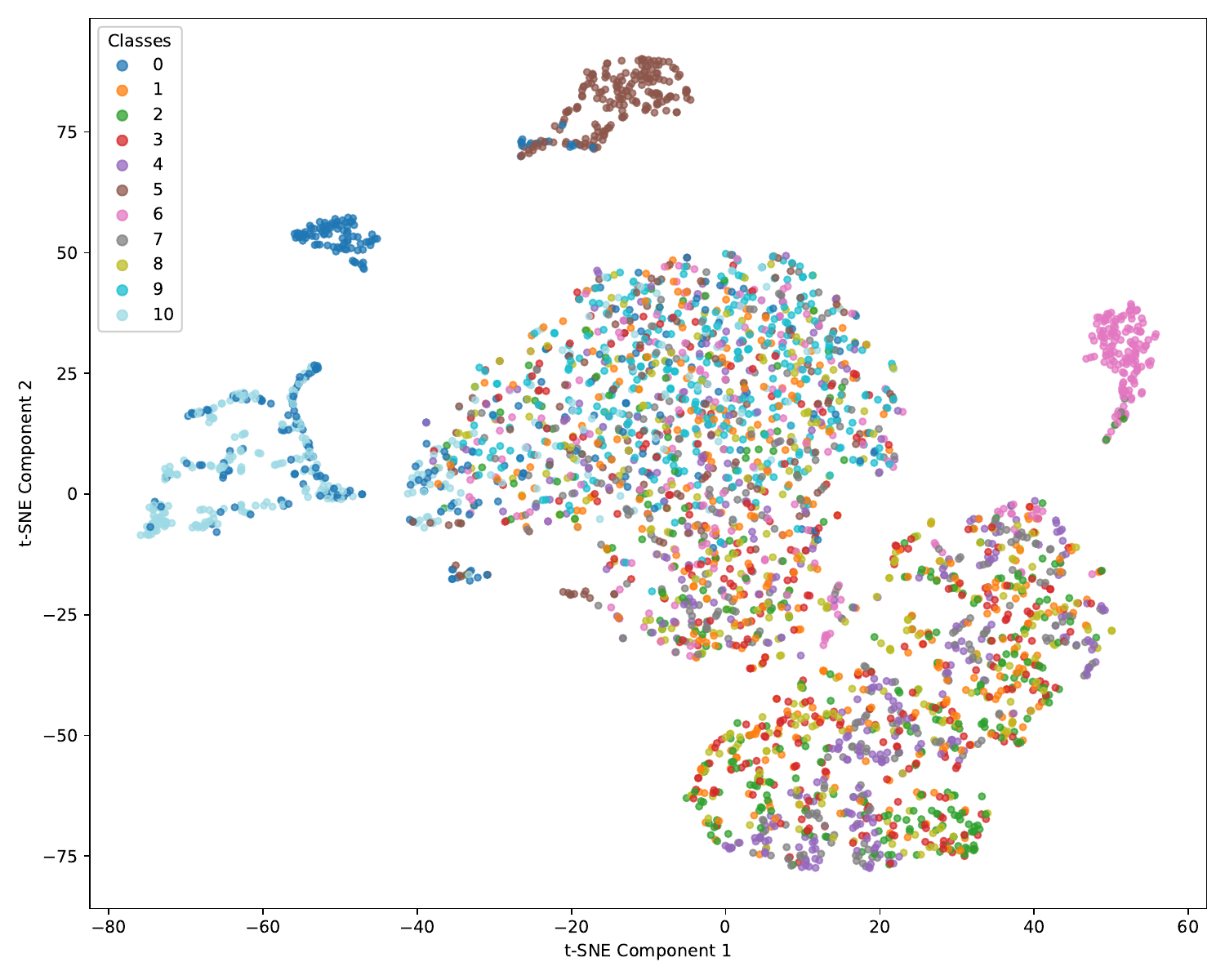}
        \subfloat{(b)}   
    \end{minipage}
    \hspace{0.00\textwidth}  
    \begin{minipage}[b]{0.24\textwidth}
        \centering
        \includegraphics[clip, trim=0.1in 0.1in 0.1in 0.1in, width=\textwidth]{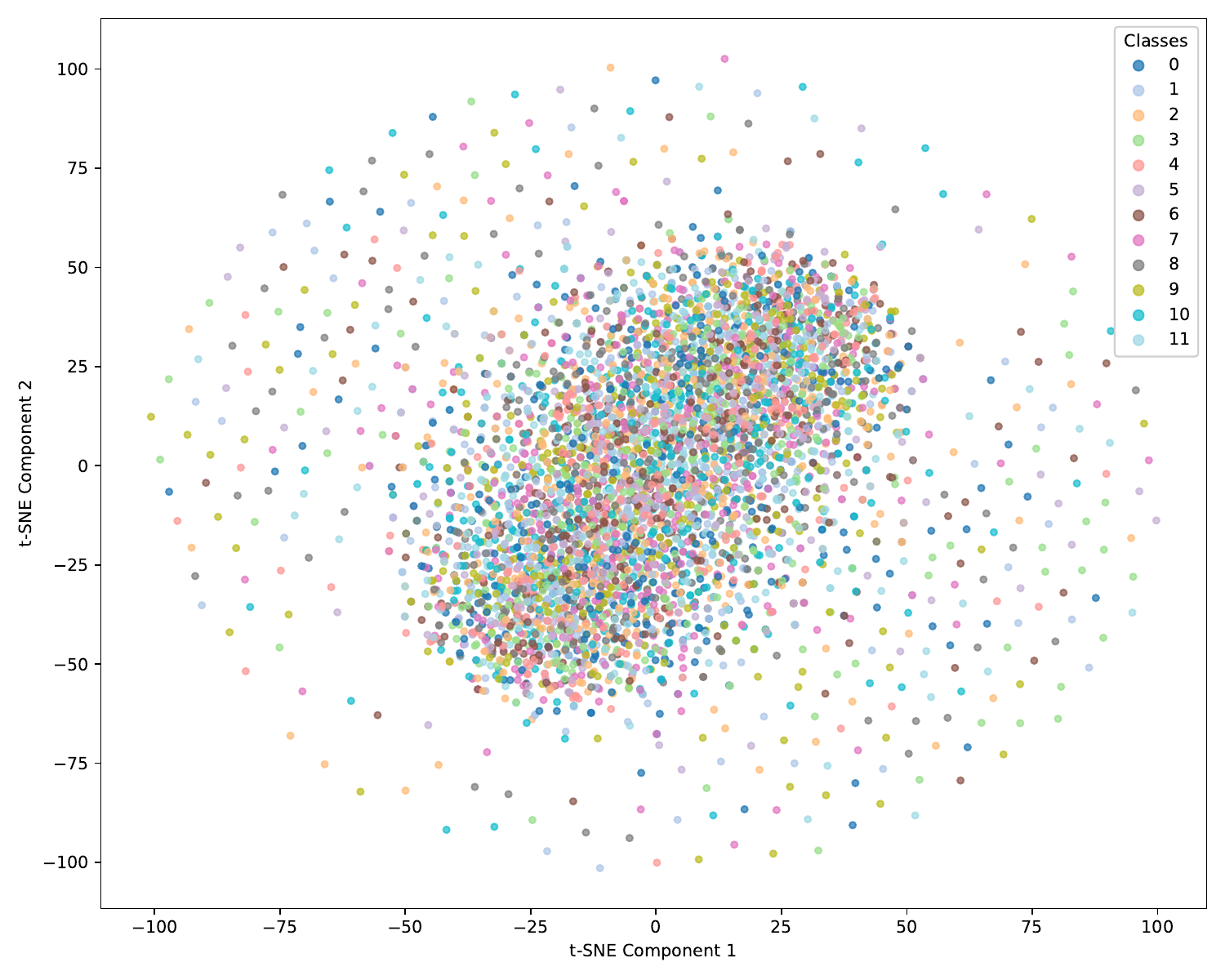}
        \subfloat{(c)}   
    \end{minipage}
    \hspace{0.00\textwidth}  
    \begin{minipage}[b]{0.24\textwidth}
        \centering
        \includegraphics[clip, trim=0.1in 0.1in 0.1in 0.1in, width=\textwidth]{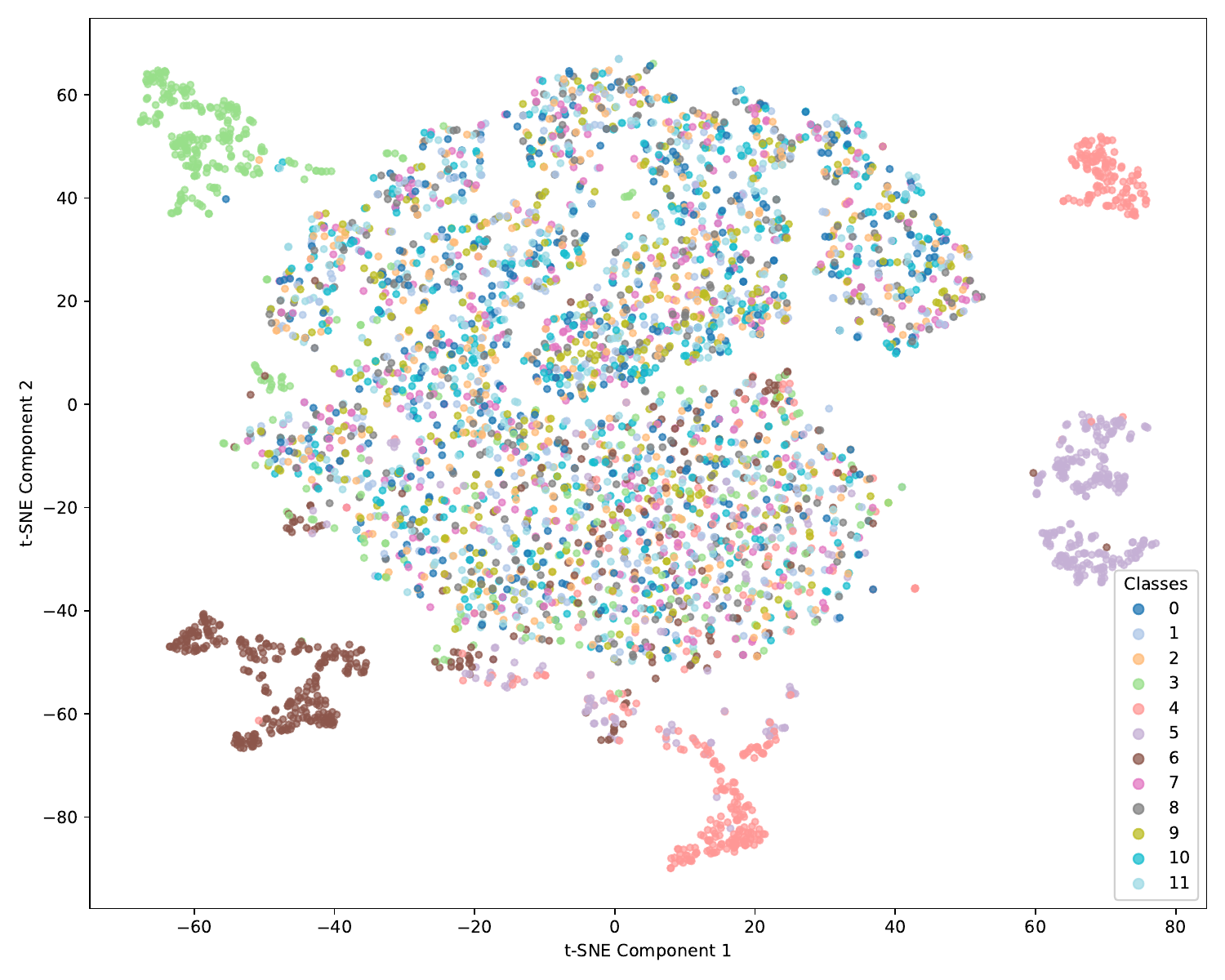}
        \subfloat{(d)}   
    \end{minipage}
    \caption{Based on the visualization of latent features from the RML2016.10a and Sig2019-12 datasets. (a) shows the original data features of RML2016.10a. (c) shows the original data features of Sig2019-12. (b) and (d) display the unsupervised features extracted by MCLRL.}
    \label{fig_7}
\end{figure*}

\textit{4) \textbf{Feature Fusion Experiment}:} We explore the performance of feature fusion methods, attention modules, and linear classifiers during the downstream phase. In the downstream phase, the three encoders generate features with consistent dimensions. The main feature fusion methods considered are addition ('A'), dot product ('D'), and direct concatenation ('C'). The attention module and linear classifier are denoted as 'AM' and 'LC', respectively. If the attention module and linear classifier are not used, the processed features are fed directly into a fully connected layer to calculate class probabilities.

The experimental results are presented in Fig.\ref{fig_6}. Using the RML2016.10a dataset as an example, the accuracy of feature fusion methods that directly add or perform a dot product on the features is much lower than that achieved by direct concatenation. The possible reason is that adding or performing a dot product on the features from the three representation domains may cause confusion and loss of feature information, while direct concatenation preserves all feature information to the fullest. When a custom attention module and linear classifier are added, the accuracy improves significantly. This is because the attention module extracts important information from features of different representation domains, and the inclusion of a Dropout layer in the linear classifier helps prevent overfitting, thus enhancing generalization, which is crucial for FSL tasks. This demonstrates the effectiveness of the attention module and linear classifier in the MCLRL framework.

\textit{5) \textbf{Encoder Ablation Experiment}:} Finally, we explore the generalization ability of the MCLRL framework, focusing on its capacity to maintain good performance across various feature extractors. Since the time-domain encoder plays a dominant role in classification features, we use the ResNet1D model and incorporate other signal classification models from Chen et al.\cite{chen2021signet}, namely CNN2D, AlexNet, and LeNet. The experiment uses the RML2016.10a dataset as an example.

The final results are presented in Table~\ref{table_4}. Overall, different time-domain encoders perform well, with accuracy depending on the feature extraction ability of the time-domain encoder itself. However, due to the inclusion of the frequency and constellation diagram domains, the final average accuracy does not change significantly. This indicates that the MCLRL framework has good generalization, as it does not excessively rely on the performance of the feature extractor. Joint learning of the three representation domains demonstrates good stability.

\subsection{Visualization}
\label{sec:Visualization}
To further validate the effectiveness of our MCLRL framework, we applied t-SNE to visualize the latent features of the RML2016.10a dataset, using it as an example. Fig.\ref{fig_7} shows the t-SNE plots of the original signal data (a) and the pretrained features of MCLRL (b). We observe that, in the high-SNR dataset, certain class features already show a high degree of clustering, further demonstrating the effectiveness of our MCLRL framework.

\section{Conclusion}
\label{sec:Conclusion}
We propose MCLRL, a novel modulation signal FSL framework, aimed at providing a more effective feature extraction scheme for modulation recognition tasks. The MCLRL framework combines multiple representation domains to enable mutual supervision learning of multi-domain features, uses reinforcement learning to select the most effective data augmentation strategies, and employs a lightweight attention module for feature fusion and classification, improving the accuracy of signal FSL. Comprehensive experimental results validate the effectiveness of MCLRL.



%

\bibliographystyle{IEEEtran}
\bibliography{references}

\begin{IEEEbiography}[{\includegraphics[width=1in,height=1.25in,clip,keepaspectratio]{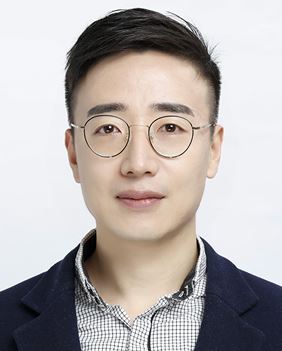}}]{Dongwei Xu}
(Member, IEEE) received the B.E. and Ph.D. degrees from the State Key Laboratory of Rail Traffic Control and Safety, Beijing Jiaotong University, Beijing, China, in 2008 and 2014, respectively. He is currently an Associate Professor with the Institute of Cyberspace Security, Zhejiang University of Technology, Hangzhou, China. His research interests include intelligent transportation Control, management, and traffic safety engineering.
\end{IEEEbiography}

\begin{IEEEbiography}[{\includegraphics[width=1in,height=1.25in,clip,keepaspectratio]{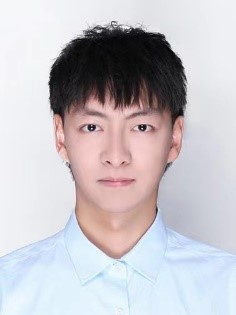}}]{Yutao Zhu}
received his bachelor's degree from the School of Electrical and Electronic Engineering at Wenzhou University in 2023. He is currently a master's student at the School of Information Engineering at Zhejiang University of Technology. His research interests include signal processing and few-shot learning.
\end{IEEEbiography} 

\begin{IEEEbiography}[{\includegraphics[width=1in,height=1.25in,clip,keepaspectratio]{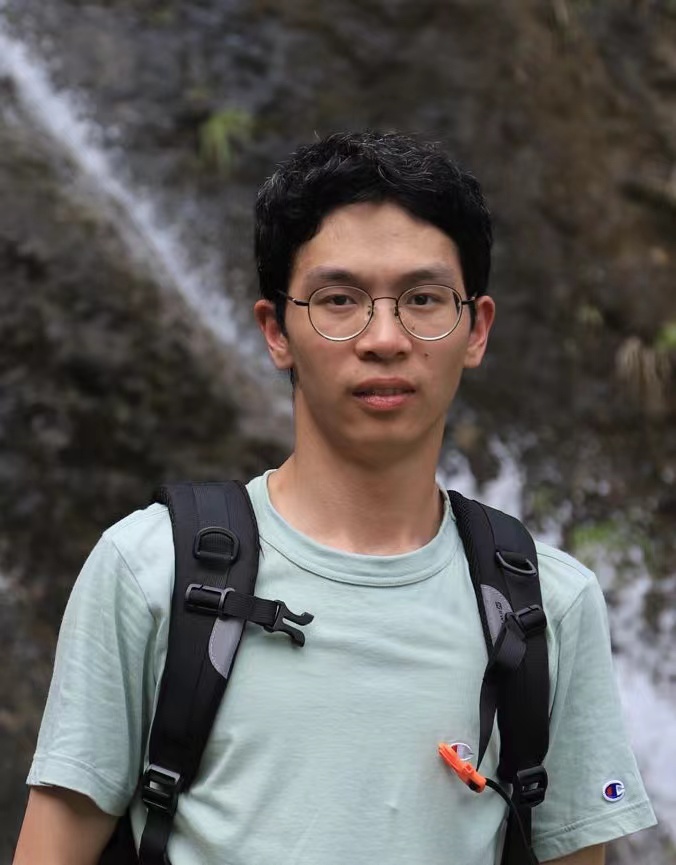}}]{Yao Lu}
received his B.S. degree from Zhejiang University of Technology and is currently pursuing a Ph.D. in control science and engineering at Zhejiang University of Technology. He has published several academic papers in international conferences and journals, including ECCV and TNNLS. His research interests include deep learning and computer vision, with a focus on explainable artificial intelligence and model compression.
\end{IEEEbiography} 

\begin{IEEEbiography}[{\includegraphics[width=1in,height=1.25in,clip,keepaspectratio]{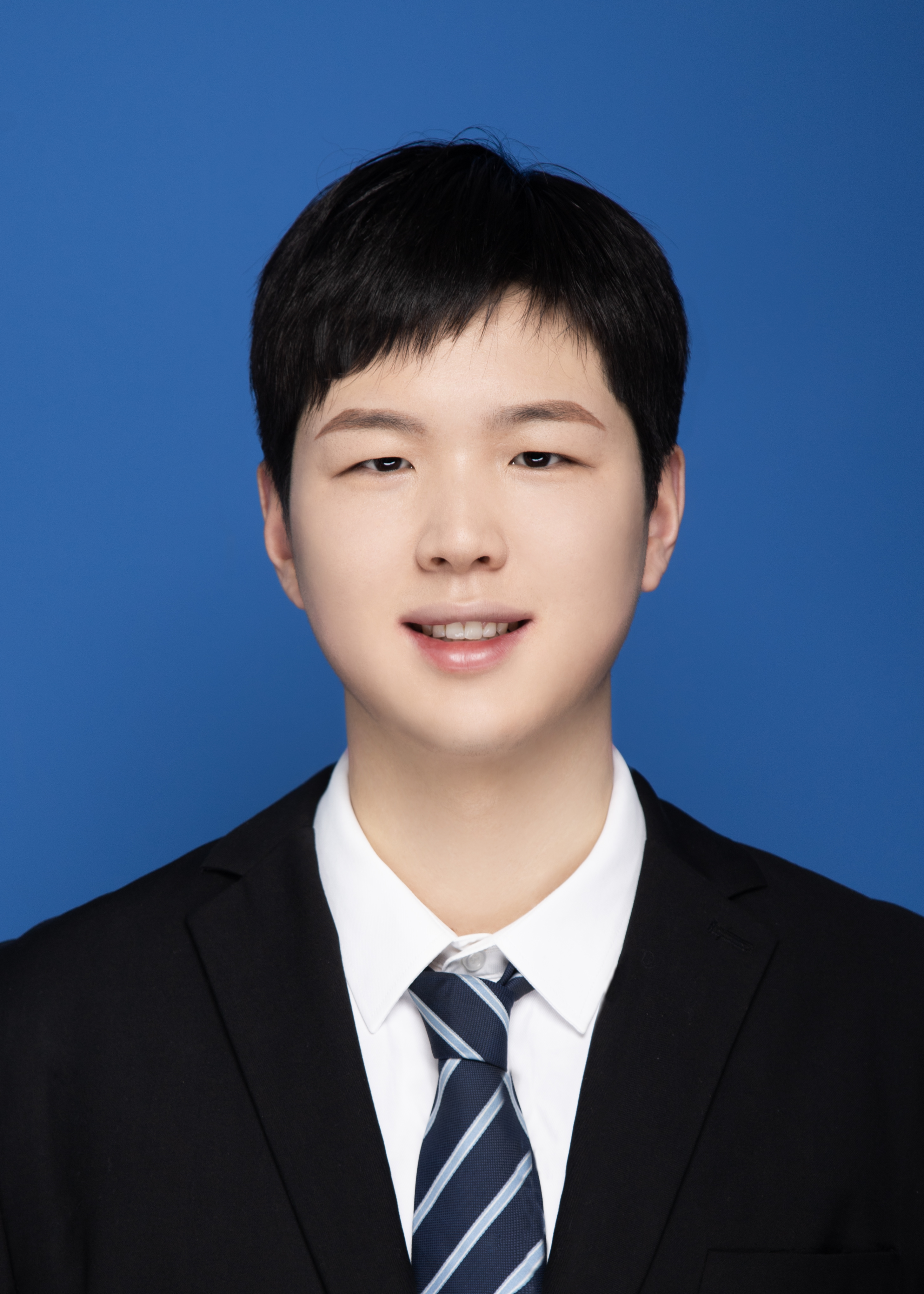}}]{Youpeng Feng}
Youpeng Feng received his Bachelor of Engineering degree from Zhengzhou Technology and Business University in 2024. He is currently pursuing a master's degree at the School of Computer Science and Technology, Zhejiang University of Technology, with research interests in automatic modulation recognition and few-shot learning.
\end{IEEEbiography} 

\begin{IEEEbiography}[{\includegraphics[width=1in,height=1.25in,clip,keepaspectratio]{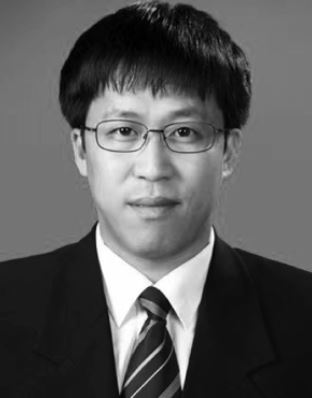}}]{Yun Lin}
(Member, IEEE) received the B.S. degree in electrical engineering from Dalian Maritime University, Dalian, China, in 2003, the M.S. degree in communication and information system from the Harbin Institute of Technology, Harbin, China, in 2005, and the Ph.D. degree in communication and information system from Harbin Engineering University, Harbin, in 2010. From 2014 to 2015, he was a Research Scholar with Wright State University, Dayton, OH, USA. He is currently a Full Professor with the College of Information and Communication Engineering, Harbin Engineering University. He has authored or coauthored more than 200 international peer-reviewed journal/conference papers, such as IEEE Transactions on Industrial Informatics, IEEE Transactions on Communications, IEEE Internet of Things Journal, IEEE Transactions on Vehicular Technology, IEEE Transactions on Cognitive Communications and Networking, TR, INFOCOM, GLOBECOM, ICC, VTC, and ICNC. His current research interests include machine learning and data analytics over wireless networks, signal processing and analysis, cognitive radio and software-defined radio, artificial intelligence, and pattern recognition.
\end{IEEEbiography}

\begin{IEEEbiography}[{\includegraphics[width=1in,height=1.25in,clip,keepaspectratio]{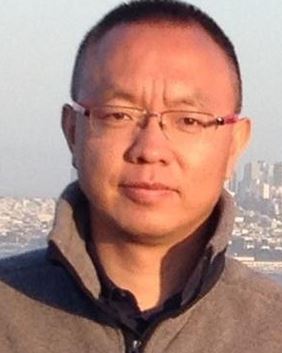}}]{Qi Xuan}
(Senior Member, IEEE) received the B.S. and Ph.D. degrees in control theory and engineering from Zhejiang University, Hangzhou, China, in 2003 and 2008, respectively. He was a Postdoctoral Researcher with the Department of Information Science and Electronic Engineering, Zhejiang University from 2008 to 2010, and a Research Assistant with the Department of Electronic Engineering, City University of Hong Kong, Hong Kong, in 2010 and 2017, respectively. From 2012 to 2014, he was a Postdoctoral Fellow with the Department of Computer Science, University of California at Davis, Davis, CA, USA. He is currently a Professor with the Institute of Cyberspace Security, College of Information Engineering, Zhejiang University of Technology, Hangzhou, and also with the PCL Research Center of Networks and Communications, Peng Cheng Laboratory, Shenzhen, China. He is also with Utron Technology Company Ltd., Xi’an, China, as a Hangzhou Qianjiang Distinguished Expert. His current research interests include network science, graph data mining, cyberspace security, machine learning, and computer vision.
\end{IEEEbiography}


\vfill

\end{document}